# Chaotic Quantum Double Delta Swarm Algorithm using Chebyshev Maps: Theoretical Foundations, Performance Analyses and Convergence Issues

**Saptarshi Sengupta *, Sanchita Basak, Richard Alan Peters II**

Department of Electrical Engineering and Computer Science, Vanderbilt University, 2201 West End Ave, Nashville, TN 37235, USA; sanchita.basak@vanderbilt.edu (S.B.); alan.peters@vanderbilt.edu (R.A.P.).

* Correspondence: saptarshi.sengupta@vanderbilt.edu; Tel.: +1 (615)-678-3419

**Abstract:** Quantum Double Delta Swarm (QDDS) Algorithm is a networked, fully-connected novel metaheuristic optimization algorithm inspired by the convergence mechanism to the center of potential generated within a single well of a spatially co-located double-delta well setup. It mimics the wave nature of candidate positions in solution spaces and draws upon quantum mechanical interpretations much like other quantum-inspired computational intelligence paradigms. In this work, we introduce a Chebyshev map driven chaotic perturbation in the optimization phase of the algorithm to diversify weights placed on contemporary and historical, socially-optimal agents' solutions. We follow this up with a characterization of solution quality on a suite of 23 single-objective functions and carry out a comparative analysis with eight other related nature-inspired approaches. By comparing solution quality and successful runs over dynamic solution ranges, insights about the nature of convergence are obtained. A two-tailed t-test establishes the statistical significance of the solution data whereas Cohen's d and Hedge's g values provide a measure of effect sizes. We trace the trajectory of the fittest pseudo-agent over all iterations to comment on the dynamics of the system and prove that the proposed algorithm is theoretically globally convergent under the assumptions adopted for proofs of other closely-related random search algorithms.

## 1. Introduction

With sensor fusion and big data taking centerstage in ubiquitous computing niches, the importance of customized, application-specific optimization paradigms is gaining recognition. The computational intelligence community is poised for exponential growth as nature-inspired modeling becomes ever more practicable in the face of abundant computational power. Thus, it is in the interest of exploratory analysis to mimic different natural systems in order to gain adequate understanding of when and on which kinds of problems certain types of biomimicry work particularly well. In this work, a subclass of the modeling paradigm of quantum-mechanical systems involving two Dirac delta potential functions is studied. The technique chosen for the study, viz. the Quantum-Double Delta Swarm (QDDS) algorithm [1] extends the well-known Quantum-behaved Particle Swarm Optimization (QPSO) [2-4] using an additional Dirac delta well and imposing motional constraints on particles to effect in convergence to a single well under the influence of both. The particles in QDDS are centrally pulled by an attractive potential field and a recursive Monte Carlo relation is established by collapse of the wavefunctions around the center of the wells. The methodology has been put forward and tested on select unimodal and multi-modal benchmarks in [1] and generates promising solution quality when compared to [4]. In this work, we primarily report performance improvements of the QDDS algorithm when its solution update process is influenced by a random perturbation drawn from a Chebyshev chaotic map. The perturbation seeks to diversify the weight array corresponding to the current and socially-optimal agents' solutions. A detailed performance characterization over twenty-three single-objective, unimodal and multi-modal functions of fixed and varying dimensions is carried out. The characterization is repeated for eight other nature-inspired approaches to provide a basis for comparison. The collective potential (cost) quality and precision data from the experimentation provide information on the operating conditions and tradeoffs while the conclusion drawn from a subsequent two-tailed t-test points to the statistical significance of the results at the $\Theta = 0.05$ level. We follow the path of the best performing agent in any iteration across all iterations and critically analyze the dynamical limitations of the algorithm (we assume that one iteration is equivalent to an atomic level function evaluation). Consequently, we also look at the global convergence proof of Random Search algorithms [5] and contend that the proposed algorithm theoretically converges to the global infimum under certain weak assumptions adopted for convergence proofs of similar random search techniques.

The organization of the article is as follows: in Section II we walk through a couple of major swarm intelligence paradigms and derive our way through the classical and quantum interpretations in these multi-agent systems. In section III, we talk about swarm propagation under the influence of a double Dirac delta well and setup its quantum mechanical model. In section IV we outline the QDDS and the Chebyshev map driven QDDS (C-QDDS) and provide an involved algorithmic procedure for purposes of reproducibility. Following this, in Section V we detail the benchmark optimization problems and graphically illustrate their three-dimensional representations. This is followed in Section VI by comparative analyses of iterations on the benchmarks and statistical significance tests, taking into account the contribution of effect sizes. The trajectory of the best performing agent in each iteration is tracked along the function contours and the limitations and successes of the approach are identified. In section VII critical analyses is presented in light of the findings. In Section VIII a global convergence proof is given for the algorithm, and finally, Section IX charts out future directions and concludes the paper.

## 2. Background

The seminal work of Eberhart and Kennedy on flocking induced stochastic, multi-particle swarming resulted in a surge in nature-inspired optimization research, specifically after their highly influential paper: Particle Swarm Optimization [6] (PSO) at the International Conference on Neural Networks in 1995. This was a landmark moment in the history of swarm intelligence and the following years saw a surge of interest towards the application of nature-inspired methods in approximating engineering problems that were till then either not tractable or simply hard from a computational standpoint. With a steady increase in processor speed and distributed computing abilities over the last couple of decades, gradient-independent approaches have gradually become ever so common. The simple and intuitive equations of motion in PSO are powerful due to simplicity and low computational cost. In this section, a formal transition from the classical model of the canonical PSO to that of quantum-inspired PSO, or the Quantum-behaved PSO (QPSO) is explored. The QPSO model assumes quantum properties in agents and establishes an uncertainty-based position distribution instead of a deterministic one as in the canonical PSO with Newtonian walks. Importantly enough, the QPSO algorithm requires the practitioner to tune only one parameter: the contraction-expansion (CE) coefficient instead of three in PSO. It is worth looking at the dynamics of a PSO-driven swarm to gain a better understanding of singular and double Dirac delta driven quantum swarms, later in the article.

### *2.1. The Classical PSO*

Assume $\boldsymbol{x_{i=1..m}} = [x_1 x_2 x_3 \ldots x_m]$ is the cohort of $m$ particles of dimensionality $n$ and $\boldsymbol{v_{i=1..m}} = [v_1 v_2 v_3 \ldots v_m]$ are the velocity vectors which denote incremental changes in their positions in the solution hyperspace. Given this knowledge, a canonical PSO-like formulation may be expressed as:

$$v_{ij}(t+1) = w \times v_{ij}(t) + C_1 \times r_1(t) \times \left(P_{ij}(t) - x_{ij}(t)\right) + C_2 \times r_2(t) \times \left(P_{gj}(t) - x_{ij}(t)\right) \tag{1}$$

$$x_{ij}(t+1) = x_{ij}(t) + v_{ij}(t+1) \tag{2}$$

The parameters $w, C_1, C_2, r_1, r_2$ are responsible for imparting inertia, cognitive and social weights as well as random perturbations towards the historical best position $P_{ij}(t)$ of any particle (*pbest*) or $P_{gj}(t)$, that of the swarm as a whole (*gbest*). The canonical PSO model mimics social information exchange in flocks of birds and schools of fish and is a simple, yet powerful optimization paradigm. However, it has its limitations: Van den Bergh showed that the algorithm is not guaranteed to converge to globally optimum solutions based on the convergence criteria put forward by Solis and Wet [5]. Clerc and Kennedy demonstrated in [7] that the algorithm may converge if particles cluster about a local attractor $p$ lying at the diagonal end of the hyper-rectangle constructed using its cognitive and social velocity vectors (terms 2 and 3 in the right-hand side of equation 1, respectively). Proper tuning of the algorithmic parameters and limits on the velocity are usually required to bring about convergent behavior. The interested reader may look at [8-11] for detailed operating conditions, possible applications and troubleshooting of issues when working with the PSO algorithm.

*2.2. The Quantum-behaved PSO*

The local attractor $p$, introduced in [7] as the point around which particles should flock in order to bring about swarm-wide convergence can be formally expressed using equation 3 and further simplifications lead to a parameter reduced form in equation 4. This result is possible of course, after the assumption that $c_1$ and $c_2$ may take on any values between 0 and 1.

$$p_{ij}(t+1) = \frac{c_1 P_{ij}(t) + c_2 P_{gj}(t)}{c_1 + c_2} \tag{3}$$

$$p_{ij}(t+1) = \varphi P_{ij}(t) + (1-\varphi) P_{gj}(t)\,, \varphi \sim U\,(0,1) \tag{4}$$

Drawing insights from this analysis, Sun et al in [2-3] outlined algorithmic working of Quantum-behaved Particle Swarm Optimization (QPSO). Instead of point representations of a particle, wavefunctions were used to provide quantitative sense about its state. The normalized probability density function **F** of a particle may be put forward as:

$$\mathbf{F}(X_{ij}(t+1)) = \frac{1}{L_{ij}(t)} exp^{\left(-2\left|p_{ij}(t) - X_{ij}(t+1)\right| / L_{ij}(t)\right)} \tag{5}$$

$L$ is the standard deviation of the distribution: it provides a measure of the dynamic range of the search space of a particle in a specific timestep. Using Monte Carlo method, equation (5) may be transformed into a recursive, computable closed form expression of particle positions in equation (6) below:

$$X_{ij}(t+1) = p_{ij}(t) \pm \frac{L_{ij}(t)}{2} \ln\left(\frac{1}{u}\right) \,, u \sim U\,(0,1) \tag{6}$$

$L$ is computed as a measure of deviation from the average of all individual personal best particle positions (*pbest*) in each dimension, i.e. the farther from the average a particle is in a dimension the larger the value of $L$ is for that dimension. This average position has been dubbed the name *'Mean Best'* or *'mbest'* and is an agglomerative representation of the swarm as if each member were in its personal best position visited in course of history.

$$mbest(t) = \left[mbest_1(t) mbest_2(t) mbest_3(t) \dots mbest_j(t)\right]$$

$$= \left[\frac{1}{m}\sum_{i=1}^{m} p_{i1}(t) \frac{1}{m}\sum_{i=1}^{m} p_{i2}(t) \frac{1}{m}\sum_{i=1}^{m} p_{i3}(t) \dots \frac{1}{m}\sum_{i=1}^{m} p_{ij}(t)\right] \tag{7}$$

Therefore, $L$ may be expressed by including the deviation from $mbest$ by equation (8). The modulation factor $\beta$ is known as the *Contraction-Expansion (CE) Factor* and may be adjusted to control the convergence speed of the QPSO algorithm depending on the application.

$$L_{ij}(t) = 2\beta\left|mbest_j(t) - X_{ij}(t)\right| \tag{8}$$

Subsequently plugging the value of $L$ obtained in equation (8) into equation (6), the position update formulation for QPSO may be re-expressed as the following:

$$X_{ij}(t+1) = p_{ij}(t) \pm \beta\left|mbest_j(t) - X_{ij}(t)\right|\ln\left(\frac{1}{u}\right) \quad , u\sim U\ (0,1) \tag{9}$$

Issues such as suboptimal convergence during the application of the QPSO algorithm may arise out of an unbiased selection of weights in the *mean best* computation as well as the overdependence on the globally best particle in the design of the local attractor $p$. These issues have also been studied by Xi et al. in [12], Sengupta et al. in [13] and Dhabal et al. [14]. Xi et al. proposed a differentially weighted *mean best* in [4]: a variant of the QPSO algorithm with a weighted mean best position (WQPSO), which seeks to alleviate the subpar selection of weights in the *mean best* update process. The underlying assumption is that fitter particles stand to contribute more to the *mean best* position and that these particles should be accorded larger weights, drawing analogy with the correlation between cultural uptick and the contributions of the societal, intellectually elite to it [4]. Xi et al. also put forward in [12] a local search strategy using a *super particle* with variable contributions from swarm members to overcome the dependence issues during the local attractor design. However, to date no significant study has been undertaken to investigate the effect of more than one spatially co-located basin of attraction around the local attractor, particularly that of multi-well systems. In the next section we seek to derive state expressions of a particle convergent upon one well under the influence of two spatially co-located Dirac-delta wells.

## 3. Swarming Under the Influence of Two Delta Potential Wells

The time-independent Schrodinger's wave equation governs the different interpretations of particle behavior:

$$[-\frac{\hbar^2}{2m}\nabla^2 + V(r)]\psi(r) = E\,\psi(r) \tag{10}$$

$\psi(r)$, $V(r)$, $m$, $E$ and $\hbar$ represent the wave function, the potential function, the reduced mass, the energy of the particle and reduced Planck's constant respectively. However, the wavefunction $\psi(r)$ has no physical significance on its own: its amplitude squared is a measure of the probability of finding a particle. Let us consider a particle under the influence of two delta potential wells experiencing an attractive potential $V$:

$$V(r) = -\mu\{\delta(r+a) + \delta(r-a)\} \tag{11}$$

The centers of the two wells are at $-a$ and $a$ and $\mu$ is a constant indicative of the depth of the wells. Under the assumption that the particle experiences no attractive potential, i.e. $V = 0$ in regions far away from the centers, the even solution of the time-independent Schrodinger's equation in equation (10) takes the following form:

$$-\frac{\hbar^2}{2m}\frac{d^2}{dr^2}\psi(r) = E\psi(r) \tag{12}$$

The even solutions to $\psi$ for $E < 0$ (bound states) in regions $\mathbb{R}1$: $r \in (-\infty, a)$, $\mathbb{R}2$: $r \in (-a, a)$ $and$ $\mathbb{R}3$: $r \in (a, \infty)$, taking k to be equal to $(\sqrt{2mE}/\hbar)$ can be expressed as has been proved in [15]:

$$\psi_{even}(r) = \begin{cases} \eta_1 exp(-kr) & r > a \\ \eta_2 exp(-kr) + \eta_3 exp(kr) & 0 < r < a \\ \eta_2 exp(kr) + \eta_3 exp(-kr) & -a < r < 0 \\ \eta_1 exp(kr) & r < -a \end{cases} \quad (13)$$

The constants $\eta_1$ and $\eta_2$ described in the above equation are obtained by: (a) solving for the continuity of the wave function $\psi_{even}$ at $r = a$ and $r = -a$ and (b) solving for the continuity of the derivative of the wave function at $r = 0$. Thus, $\psi_{even}$ may be re-written below as has been in [15]:

$$\psi_{even}(r) = \begin{cases} \eta_2\{1 + \exp(2ka)\}\exp(-kr) & r > a \\ \eta_2\{\exp(-kr) + \exp(kr)\} & -a < r < a \\ \eta_2\{1 + \exp(2ka)\}\exp(kr) & r < -a \end{cases} \quad (14)$$

The odd wave function $\psi_{odd}$ does not guarantee that a solution would be found [15]. Additionally, the bound state energy in double well setup is lower than that in a single well setup by approximately a factor of $(1.11)^2 \approx 1.2321$ [16]:

$$E_{bs,Double\ Well} = -(1.11)^2 E_{bs,Single\ Well} \quad (15)$$

To study the motional aspect of a particle its probability density function given by the squared magnitude of $\psi_{even}$ is formally expressed. Further, the claim that there is greater than 50% probability of a particle existing in neighborhood of the center of any of the potential wells (assumed centered at 0) boils down to the following criterion being met [2]:

$$\int_{-|r|}^{|r|} \psi_{even}(r)^2\, dr > 0.5 \quad (16)$$

$-|r|$ and $|r|$ are the dynamic limits of the neighborhood. Doing away with the inequality, equation (16) is re-written as:

$$\int_{-|r|}^{|r|} \psi_{even}(r)^2\, dr = 0.5\lambda \qquad (1 < \lambda < 2) \quad (17)$$

Equation (17) is the criterion for localization around the center of a potential well in a double Dirac delta well:

## 4. The Quantum Double Delta Swarm (QDDS) Algorithm

To ease computations, we make the assumption that one of the two potential wells is centered at 0. Then, solving for conditions of localization of the particle in the neighborhood around the center of that well and computing $\int_{-|r|}^{|r|}(\psi(r)^2\, dr$ for regions $\mathbb{R}2_{0-}$: $r' \in (-r, 0)$ and $\mathbb{R}2_{0+}$: $r' \in (0, r)$, we obtain the relationship below:

$$\eta_2{}^2 = \frac{k\lambda}{\exp(2kr) - 5\exp(-2kr) + 4kr + 4} \quad (18)$$

Replacing denominator of R.H.S. of equation (18) i.e. $(\exp(2kr) - 5\exp(-2kr) + 4kr + 4)$ as δ, we re-write it as:

$$\delta = \exp(2kr) - 5\exp(-2kr) + 4kr + 4 \quad (19)$$

Equating $B^2$ in L.H.S. of equation (18) for any two consecutive iterations (assuming it is a constant over iterations as it not a function of time) we get equations (20), (21) and (22):

$$\frac{\lambda_t}{\exp(2kr_t) - 5\exp(-2kr_t) + 4kr_t + 4} = \frac{\lambda_{t-1}}{\exp(2kr_{t-1}) - 5\exp(-2kr_{t-1}) + 4kr_{t-1} + 4} \quad (20)$$

$$\Rightarrow \frac{\lambda_t}{\delta_t} = \frac{\lambda_{t-1}}{\delta_{t-1}} \tag{21}$$

$$\Rightarrow \delta_t = \Lambda \,.\, \delta_{t-1} \qquad (0.5 < \Lambda < 2) \tag{22}$$

$\Lambda$ is the ratio $(\lambda_t/\lambda_{t-1})$ and it may vary between 0.5 to 2 since (1< $\lambda$ <2). To keep a particle constrained within the vicinity of the center of the potential well, it must meet the following condition:

$$\frac{1}{2}\delta_{t-1} < \delta_t < 2\,\delta_{t-1} \tag{23}$$

Thus, we find $\delta_t$ for any iteration by utilizing $\delta_{t-1}$, obtained in the immediately past iteration. This is done by accounting for a correction factor in the form of the gradient of $\delta_{t-1}$, multiplied by a learning rate $\alpha$. The computation of $\delta_t$ from $\delta_{t-1}$ feeds off the relationship of $\delta_{t-1}$ with $\delta_{t-2}$ while taking the sign of the gradient of $\delta_{t-1}$ into consideration. The procedural details are outlined in Algorithm 1. The learning rate $\alpha$ is chosen as a linearly decreasing, time-varying one (LTV) to help facilitate exploration of the solution space early on in the optimization phase and a gradual shift to exploitation as the process evolves. $\nu$ is a small fraction between 0 and 1 chosen at will. However, one empirically successful value is 0.3 and we use it in our computations.

$$\alpha = (1-\nu)\left(\frac{maximum\ number\ of\ \ iterations - current\ iteration}{maximum\ number\ of\ \ iterations}\right) + \nu \tag{24}$$

Upon computing a value for $\delta_t$, equation (19) is solved to retrieve an estimate of $r_t$, which denotes a candidate position as well as a potential solution at the end of that iteration.

$$r_t \cong Solve[\{\ \delta - (\exp(2kr) - \ 5\exp(-2kr) + 4kr + 4)\} = 0] \tag{25}$$

We let $r_t$ i.e. a particle's position in the current iteration maintain a component towards the best position found so far *(gbest)* in addition to its current solution obtained from equation (19). Let $\rho$ denote the component towards the *gbest* position and (1- $\rho$) be that towards the current solution.

$$r_t^{new} = \rho r_t + (1-\rho) r_{gbest} \tag{26}$$

A cost function is subsequently computed and the corresponding particle position is saved if the cost is lowest among all the historical swarm-wide best costs obtained. This process is repeated until convergence criteria of choice (solution accuracy threshold, computational expense, memory requirements, success rate etc.) are met.

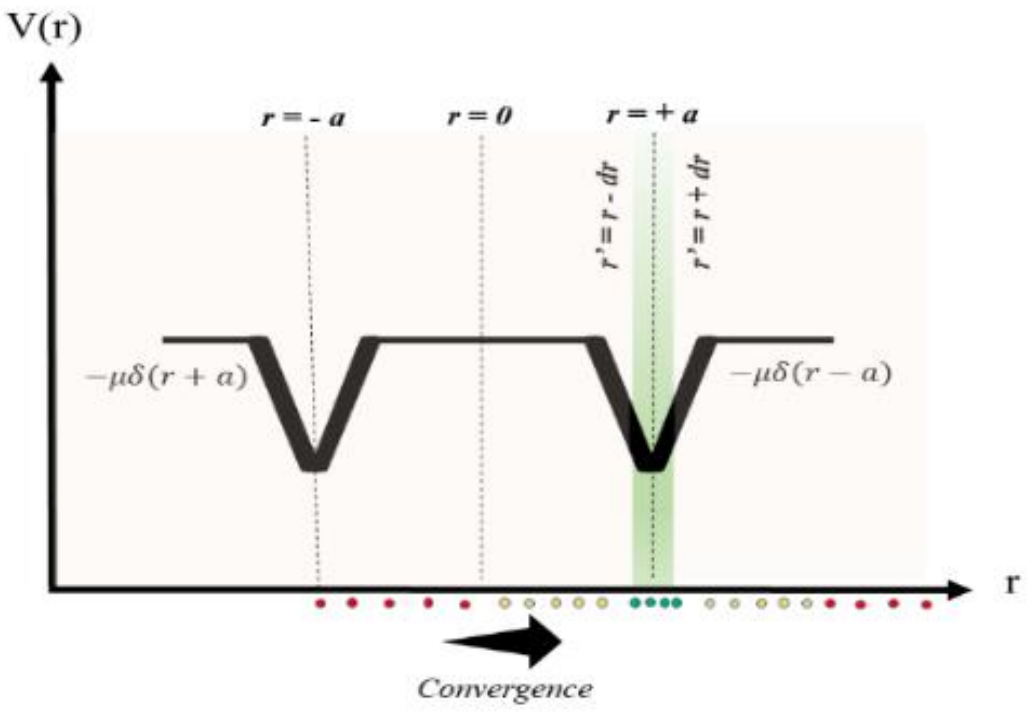

Figure 1: The Double Well Potential Setup

### 4.1 QDDS with Chaotic Chebyshev Map (C-QDDS)

In this section, we use a Chebyshev chaotic map to generate co-efficient sequences for driving the belief $\rho$ in the solution update phase of the QDDS algorithm.

*4.1.1. Chebyshev Map Driven Solution Update*

*Motivation:*

Chaotic metaheuristics necessitate control over the balance between diversification and intensification phases. The diversification phase is carried out by choosing an appropriate chaotic system which performs the extensive search while the intensification phase is carried out by performing a local search such as gradient descent. It is important that during the initial progression of the search, multiple orbits pass through the vicinity of the local extrema. A large perturbation weight ensures that the strange attractor of one local extremum intersects the strange attractor of any of the other local extrema [31]. To this end, we generate a sorted sequence which acts as a perturbation source of tapering magnitude using the Chebyshev chaotic map using the recursive relation in Equation (27) [17]. There is a relative dearth of studies looking at chaotic perturbations to agent positions to drive them towards socially optimal agent locations. In our approach, we look to facilitate extensive communication among agents by employing larger chaotic weights (diversification phase) in the initial stages and local communication among agents by tapering weights (intensification phase) with the progression of function evaluations. The optimal choice and arrangement of the modulus and sign of the weights generated using the pseudo random number generator or any other method for that matter is subject to change with a change in the application problem and is very much an open question in exploration-exploitation based search niche. However, the two properties of ergodicity and non-repetition in chaotic time sequences have proved useful in a number of related classical studies [32-34] and are key factors supporting the choice of the perturbation weights in this work. Furthermore, the properties of large Lyapunov co-efficient (a measure of chaoticity) and space-filling nature of the Chebyshev sequence serve to help avoid stagnation in local extrema and supplement the choice of the type of chaotic map in the studies in this article.

$$\rho_t^{Chebyshev} = \cos\left(t * cos^{-1}\left(\rho_{t-1}^{Chebyshev}\right)\right) \quad (27)$$

Equation (26) subsequently becomes:

$$r_t^{new} = \rho_t^{Chebyshev} * r_{iter} + \left(1 - \rho_t^{Chebyshev}\right) * r_{gbest} \quad (28)$$

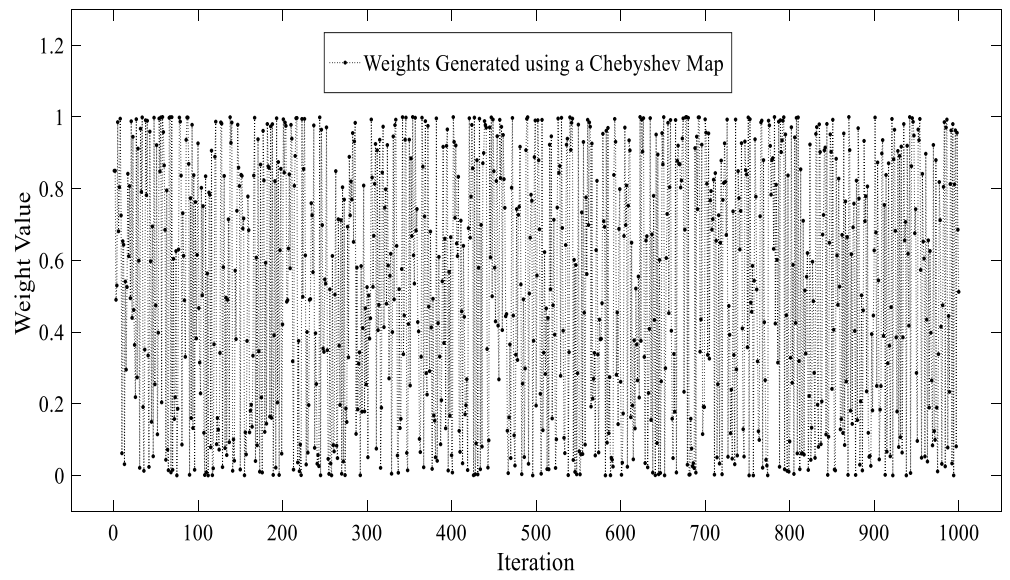

Figure 2: Generated Weights ($\rho^{Chebyshev}$) from a Chebyshev Chaotic Map over 1000 Iterations

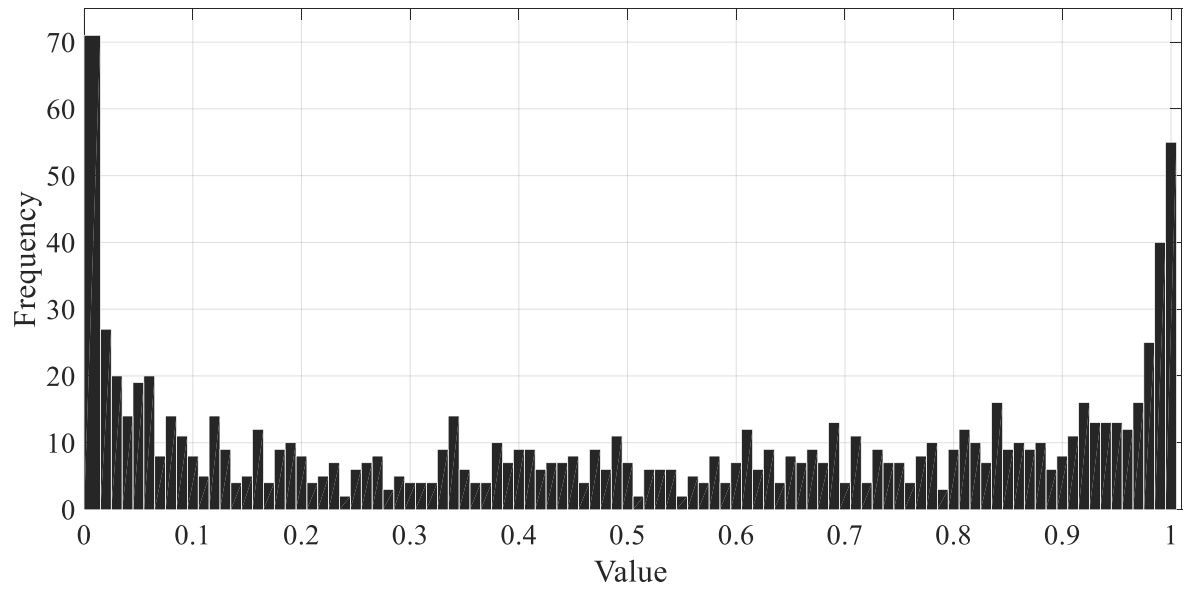

Figure 3: Histogram of Generated Weights ($\rho^{Chebyshev}$) from the Chebyshev Map over 1000 Iterations

**Table 1. General Terms used in Context of the Algorithms and Experimentation**

| Term | Discussion |
|---|---|
| *Some General Terms* | |
| **Population (X)** | The collection or *'swarm'* of agents employed in the search space |
| **Fitness Function (f)** | A measure of convergence efficiency |
| **Current Iteration** | The ongoing iteration among a batch of dependent/independent runs |
| **Maximum Iteration Count** | The maximum number of times runs are to be performed |
| *Particle Swarm Optimization (PSO)* | |
| **Position (X)** | Position value of individual swarm member in multidimensional space |
| **Velocity (v)** | Velocity values of individual swarm members |
| **Cognitive Accl. Coefficient (C1)** | Empirically found scale factor of pBest attractor |
| **Social Accl. Co-efficient (C2)** | Empirically found scale factor of gBest attractor |
| **Personal Best (pBest)** | Position corresponding to historically best fitness for a swarm member |
| **Global Best (gBest)** | Position corresponding to best fitness over history for swarm members |
| **Inertia Weight Co-efficient (ω)** | Facilitates and modulates exploration in the search space |
| **Cognitive Random Perturbation ($r_1$)** | Random noise injector in the Personal Best attractor |
| **Social Random Perturbation ($r_2$)** | Random noise injector in the Global Best attractor |
| *Quantum-behaved Particle Swarm Optimization (QPSO)* | |
| **Local Attractor** | Set of local attractors in all dimensions |
| **Characteristic Length** | Measure of scales on which significant variations occur |
| **Contraction-Expansion Parameter (β)** | Scale factor influencing the convergence speed of QPSO |
| **Mean Best** | Mean of personal bests across all particles, akin to leader election in species |
| *Quantum Double-Delta Swarm Optimization (QDDS)* | |
| $\boldsymbol{\rho}$ | Component towards the global best position *gbest* |
| $\boldsymbol{\psi}(\boldsymbol{r})$ | Wavefunction in the Schrodinger's equation |
| $\boldsymbol{\psi}_{\boldsymbol{even}}(\boldsymbol{r})$ | Even solutions to the Schrodinger's Equation for a Double Delta Potential Well |
| $\boldsymbol{V}(\boldsymbol{r})$ | Potential Function |
| $\boldsymbol{\Lambda}$ | Limiter |
| $\boldsymbol{\delta}_{\boldsymbol{iter}}$ | Characteristic Constraint |
| $\epsilon$ | A small fraction between 0 and 1 chosen at will |
| $\mathbb{R}\mathbf{1}$: $\boldsymbol{r} \in (-\infty, \boldsymbol{a})$ | Region 1 |
| $\mathbb{R}\mathbf{2}$: $\boldsymbol{r} \in (-\boldsymbol{a}, \boldsymbol{a})$ | Region 2 |
| $\mathbb{R}\mathbf{3}$: $\boldsymbol{r} \in (\boldsymbol{a}, \infty)$ | Region 3 |
| $\boldsymbol{\alpha}$ | Learning Rate |
| $\boldsymbol{\rho}_{iter}^{Chebyshev}$ | Component towards the global best position *gbest* drawn from a Chebyshev map |
| $\boldsymbol{\mu}$ | Depth of the wells |
| $\boldsymbol{a}$ | Co-ordinate of wells |

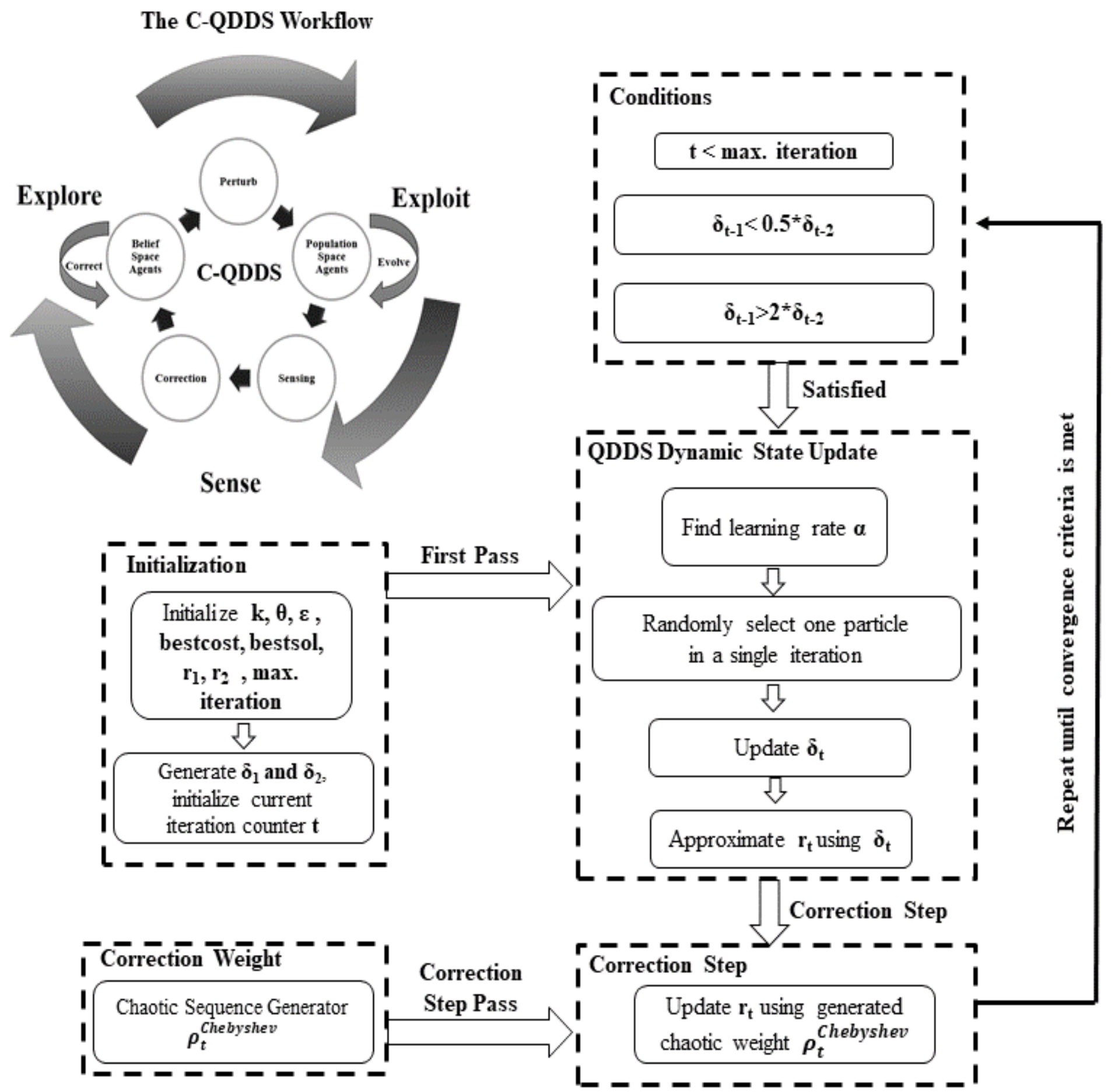

Figure 4: Schematic of the C-QDDS Workflow

### *4.1.2. Pseudocode of the C-QDDS Algorithm*

In this section, we present the pseudocode of the Chaotic Quantum Double Delta Swarm (C-QDDS) Algorithm.

**Algorithm 1.** Quantum Double Delta Swarm Algorithm

***Initialization Phase***

*Initialize* $\mathbf{k}$
*Initialize scale factor* $\theta$ *randomly* ($\approx 10^{-3}$)
*Initialize a constant* $\varepsilon$ *between 0 and 1 as the lower bound of* $\chi$
*Initialize maximum number of iterations as* **max. iterations**
*Initialize the global best cost as* **bestcost** *and global best position as* **bestsol**
*for each particle*
    *for each dimension*
        *Initialize positions* $\mathbf{r_1}$ *and* $\mathbf{r_2}$ *for iterations 1 and 2*
    *end for*
*end for*
*Generate* $\boldsymbol{\delta_1}$ *and* $\boldsymbol{\delta_2}$ *from* $\mathbf{r_1}$ *and* $\mathbf{r_2}$ *using equation (19)*
*Set current iteration* $t$ =3

***Optimization Phase***

***while*** (**t** < **max. iterations**) *and* {($\boldsymbol{\delta_{t-1}} < \mathbf{0.5} * \boldsymbol{\delta_{t-2}}$) *or* ($\boldsymbol{\delta_{t-1}} > \mathbf{2} * \boldsymbol{\delta_{t-2}}$)}
    *Find learning rate* $\boldsymbol{\alpha}$ *using eq. (24)*
    *Select a particle randomly*
    ***for*** *each dimension*
        ***if*** ($\boldsymbol{\delta_{t-1}} > \mathbf{2} * \boldsymbol{\delta_{t-2}}$) *and* $\nabla\boldsymbol{\delta_{t-1}}$>0
            $\boldsymbol{\delta_t} = \boldsymbol{\delta_{t-1}} - \boldsymbol{\theta} * \nabla\boldsymbol{\delta_{t-1}} * \boldsymbol{\alpha}$
        ***else if*** ($\boldsymbol{\delta_{t-1}} > \mathbf{2} * \boldsymbol{\delta_{t-2}}$) *and* $\nabla\boldsymbol{\delta_{t-1}}$<0
            $\boldsymbol{\delta_t} = \boldsymbol{\delta_{t-1}} + \boldsymbol{\theta} * \nabla\boldsymbol{\delta_{t-1}} * \boldsymbol{\alpha}$
        ***else if*** ($\boldsymbol{\delta_{t-1}} < \mathbf{0.5} * \boldsymbol{\delta_{t-2}}$) *and* $\nabla\boldsymbol{\delta_{t-1}}$<0
            $\boldsymbol{\delta_t} = \boldsymbol{\delta_{t-1}} - \boldsymbol{\theta} * \nabla\boldsymbol{\delta_{t-1}} * \boldsymbol{\alpha}$
        ***else if*** ($\boldsymbol{\delta_{t-1}} < \mathbf{0.5} * \boldsymbol{\delta_{t-2}}$) *and* $\nabla\boldsymbol{\delta_{t-1}}$>0
            $\boldsymbol{\delta_t} = \boldsymbol{\delta_{t-1}} + \boldsymbol{\theta} * \nabla\boldsymbol{\delta_{t-1}} * \boldsymbol{\alpha}$
        ***end if***
    ***end for***
    *Solve* $\mathbf{r_t}$ *from* $\boldsymbol{\delta_t}$

    ***Chaotic Random Number Generation and Correction Phase***

    *Generate* $\boldsymbol{\rho} \in [\mathbf{0}, \mathbf{1}]$ *using Chebyshev recurrence:* $\boldsymbol{\rho}_t^{\mathbf{Chebyshev}} = \mathbf{cos}\left(\mathbf{t} * \mathbf{cos}^{-1}\left(\boldsymbol{\rho}_{t-1}^{\mathbf{Chebyshev}}\right)\right)$
    $\mathbf{r}_t^{(\mathbf{updated})} = \left(\boldsymbol{\rho}_t^{\mathbf{Chebyshev}}\right)\mathbf{r_t} + \left(\mathbf{1} - \left(\boldsymbol{\rho}_t^{\mathbf{Chebyshev}}\right)\right)\mathbf{r_{gbest}}$
    *Compute cost using* $\mathbf{r}_t^{(\mathbf{updated})}$
    ***if*** $\mathbf{cost_t}$<**bestcost**
        **bestcost** = $\mathbf{cost_t}$
        **bestsol** = $\mathbf{r}_t^{(\mathbf{updated})}$
    ***end if***
    **t** = **t** +1
***end while***

## 5. Experimental Setup

### *5.1. Benchmark Functions*

A suite of the following 23 optimization benchmark functions (F1-F23) are popularly used to inspect the performance of evolutionary optimization paradigms and have been utilized in this work to characterize the behavior of C-QDDS across unimodal and multimodal function landscapes of fixed and varying dimensionality.

**Table 2. Unimodal Test Functions Considered for Testing**

| Number | Name | Expression | Range | Min |
|---|---|---|---|---|
| F1 | Sphere | $f(x) = \sum_{i=1}^{n} x_i^2$ | [-100,100] | f(x*) = 0 |
| F2 | Schwefel's Problem 2.22 | $f(x) = \sum_{i=1}^{n} \lvert x_i \rvert + \prod_{i=1}^{n} \lvert x_i \rvert$ | [-10,10] | f(x*) = 0 |
| F3 | Schwefel's Problem 1.2 | $f(x) = \sum_{i=1}^{n} \left( \sum_{j=1}^{i} x_j \right)^2$ | [-100,100] | f(x*) = 0 |
| F4 | Schwefel's Problem 2.21 | $f(x) = max_i\{\lvert x_i \rvert, 1 \leq i \leq n\}$ | [-100,100] | f(x*) = 0 |
| F5 | Generalized Rosenbrock's Function | $f(x) = \sum_{i=1}^{n-1} [100(x_{i+1} - x_i^2)^2 + (x_i - 1)^2]$ | [-n,n] | f(x*) = 0 |
| F6 | Step Function | $f(x) = \sum_{i=1}^{n} (\lfloor x_i + 0.5 \rfloor)^2$ | [-100,100] | f(x*) = 0 |
| F7 | Quartic Function i.e. Noise | $f(x) = \sum_{i=1}^{n} ix^4 + random[0,1)$ | [-1.28,1.28] | f(x*) = 0 |

**Table 3. Multimodal Test Functions Considered for Testing**

| Number | Name | Expression | Range | Min |
|---|---|---|---|---|
| F8 | Generalized Schwefel's Problem 2.26 | $f(x) = -\sum_{i=1}^{n} (x_i \sin(\sqrt{\lvert x_i \rvert}))$ | [-500,500] | f(x*) = -12569.5 |
| F9 | Generalized Rastrigrin's Function | $f(x) = An + \sum_{i=1}^{n} [x_i{}^2 - Acos(2\pi x_i)]$ , A=10 | [-5.12, 5.12] | f(x*) = 0 |
| F10 | Ackley's Function | $f(x) = -20 \exp\left( -0.2 \sqrt{\frac{1}{d} \sum_{i=1}^{d} x_i^2} \right) - \exp\left( \sqrt{\frac{1}{d} \sum_{i=1}^{d} \cos(2\pi x_i)} \right) + 20 + \exp(1)$ | [-32.768,32.768] | f(x*) = 0 |
| F11 | Generalized Griewank Function | $f(x) = 1 + \frac{1}{4000} \sum_{i=1}^{n} x_i^2 - \prod_{i=1}^{n} \cos(\frac{x_i}{\sqrt{i}})$ | [-600,600] | f(x*) = 0 |
| F12 | Generalized Penalized Function 1 | $f(x) = \frac{\pi}{d} \left\{ 10sin^2(\pi y_1) + \sum_{i=1}^{n-1} (y_i - 1)^2 [1 + 10sin^2(\pi y_{i+1}) + (y_n - 1)^2] + \sum_{i=1}^{n} u(x_i, 10,100,4) \right\}$ | [-50,50] | f(x*) = 0 |
| F13 | Generalized Penalized Function 2 | $f(x) = 0.1\left\{ \begin{array}{c} \sin^2(3\pi x_1) \\ + \sum_{i=1}^{n-1}(x_i - 1)^2[1 + \sin^2(3\pi x_{i+1})] + (x_n - 1)^2[1 + \sin^2(2\pi x_{30})] \\ + \sum_{i=1}^{n} u(x_i, 5,100,4) \end{array} \right\}$ where $u(x_i, 5,100,4) = \begin{cases} k(x_i - a)^m, & x_i > a \\ 0, & -a < x_i < a \\ k(-x_i - a)^m, & x_i < -a \end{cases}$ $y_i = 1 + \frac{1}{4}(x_i + 1)$ | [-50,50] | f(x*) = 0 |

**Table 4. Multimodal Test Functions with Fixed Dimensions Considered for Testing**

| Number | Name | Expression | Range | Min |
|---|---|---|---|---|
| F14, n=2 | Shekel's Foxholes Function | $f(x) = \left[\frac{1}{500} + \sum_{j=1}^{25} \frac{1}{j + \sum_{i=1}^{2}(x_i - a_{ij})^6}\right]^{-1}$ where $a_{ij} = \begin{pmatrix} -32 & -16 & 0 & 16 & 32 & -32 & \dots & 0 & 16 & 32 \\ -32 & -32 & -32 & -32 & -32 & -16 & \dots & 32 & 32 & 32 \end{pmatrix}$ | [-65.536, 65.536] | f(x*) ≈ 1 |
| F15, n=4 | Kowalik's Function | $f(x) = \left[\sum_{i=1}^{11} a_i - \frac{x_1(b_i{}^2 + b_i x_2)}{b_i{}^2 + b_i x_3 + x_4}\right]^2$ Co-efficients are defined according to Table F15. | [-5,5] | f(x*) ≈ 0.0003075 |
| F16, n=2 | Six-Hump Camel-Back Function | $f(x) = 4x_1{}^2 - 2.1x_1{}^4 + \frac{1}{3}x_1{}^6 + x_1x_2 - 4x_2{}^2 + 4x_2{}^4$ | [-5,5] | f(x*) = -1.0316285 |
| F17, n=2 | Branin Function | $f(x) = \left(x_2 - \frac{5.1}{4\pi^2}x_1{}^2 + \frac{5}{\pi}x_1 - 6\right)^2 + 10\left(1 - \frac{1}{8\pi}\right)\cos x_1 + 10$ | $-5 \le x_1 \le 10$, $0 \le x_2 \le 15$ | f(x*) =0.398 |
| F18, n=2 | Goldstein-Price Function | $f(x) = [1 + (x_1 + x_2 + 1)^2(19 - 14x_1 + 3x_1{}^2 - 14x_2 + 6x_1x_2 + 3x_2{}^2)][30 + (2x_1 - 3x_2)^2(18 - 32x_1 + 12x_1{}^2 + 48x_2 - 36x_1x_2 + 27x_2{}^2)]$ | [-2,2] | f(x*) =3 |
| F19, n=3 | Hartman's Family Function 1 | $f(x) = -\sum_{i=1}^{4} c_i \exp\left[-\sum_{j=1}^{3} a_{ij}\,(x_j - p_{ij})^2\right]$ | $0 \le x_j \le 1$ | f(x*) =-3.86 |
| F20, n=6 | Hartman's Family Function 2 | $f(x) = -\sum_{i=1}^{4} c_i \exp\left[-\sum_{j=1}^{6} a_{ij}\,(x_j - p_{ij})^2\right]$ Co-efficients are defined according to Table F20.1 and F20.2 respectively. | $0 \le x_j \le 1$ | f(x*) =-3.86 |
| F21, n=4 | Shekel's Family Function 1 | $f(x) = -\sum_{i=1}^{5}[(x - a_i)(x - a_i)^T + c_i]^{-1}$ Co-efficients are defined according to Table F21. | $0 \le x_j \le 10$ | $f(x^*_{local}) = \frac{1}{c_i}$, $1 \le i \le m$ |
| F22, n=4 | Shekel's Family Function 2 | $f(x) = -\sum_{i=1}^{7}[(x - a_i)(x - a_i)^T + c_i]^{-1}$ Co-efficients are defined according to Table F22. | $0 \le x_j \le 10$ | $f(x^*_{local}) = \frac{1}{c_i}$, $1 \le i \le m$ |
| F23, n=4 | Shekel's Family Function 3 | $f(x) = -\sum_{i=1}^{10}[(x - a_i)(x - a_i)^T + c_i]^{-1}$ Co-efficients are defined according to Table F23. | $0 \le x_j \le 10$ | $f(x^*_{local}) = \frac{1}{c_i}$, $1 \le i \le m$ |

### Table 5. Co-efficients of Kowalik's Function (F15)

| Index (i) | $a_i$ | $a_{ij}^{-1}$ |
|---|---|---|
| 1 | 0.1957 | 0.25 |
| 2 | 0.1947 | 0.5 |
| 3 | 0.1735 | 1 |
| 4 | 0.1600 | 2 |
| 5 | 0.0844 | 4 |
| 6 | 0.0627 | 6 |
| 7 | 0.0456 | 8 |
| 8 | 0.0342 | 10 |
| 9 | 0.0323 | 12 |
| 10 | 0.0235 | 14 |
| 11 | 0.0246 | 16 |

### Table 6. Co-efficients of Hartman's Functions (F19)

| Index (i) | $a_{ij}, j = 1, 2, 3$ | | | $c_i$ | $p_{ij}, j = 1, 2, 3$ | | |
|---|---|---|---|---|---|---|---|
| 1 | 3 | 10 | 30 | 1 | 0.3689 | 0.1170 | 0.2673 |
| 2 | 0.1 | 10 | 35 | 1.2 | 0.4699 | 0.4387 | 0.7470 |
| 3 | 3 | 10 | 30 | 3 | 0.1091 | 0.8732 | 0.5547 |
| 4 | 0.1 | 10 | 35 | 3.2 | 0.038150 | 0.5743 | 0.8828 |

### Table 7. Co-efficients of Hartman's Functions (F20)

| Index (i) | $a_{ij}, j = 1, 2, 3$ | | | | | | $c_i$ | $p_{ij}, j = 1, 2, 3$ | | | | | |
|---|---|---|---|---|---|---|---|---|---|---|---|---|---|
| 1 | 10 | 3 | 17 | 3.5 | 1.7 | 8 | 1 | 0.1312 | 0.1696 | 0.5569 | 0.0124 | 0.8283 | 0.5886 |
| 2 | 0.5 | 10 | 17 | 0.1 | 8 | 14 | 1.2 | 0.2329 | 0.4135 | 0.8307 | 0.3736 | 0.1004 | 0.9991 |
| 3 | 3 | 3.5 | 1.7 | 10 | 17 | 8 | 3 | 0.2348 | 0.1415 | 0.3522 | 0.2883 | 0.3047 | 0.6650 |
| 4 | 17 | 8 | 0.05 | 10 | 0.1 | 14 | 3.2 | 0.4047 | 0.8828 | 0.8732 | 0.5743 | 0.1091 | 0.0381 |

### Table 8. Co-efficients of Shekel's Functions (F21-F23)

| Index (i) | $a_{ij}, j = 1, \ldots, 4$ | | | | $c_i$ |
|---|---|---|---|---|---|
| 1 | 4 | 4 | 4 | 4 | 0.1 |
| 2 | 1 | 1 | 1 | 1 | 0.2 |
| 3 | 8 | 8 | 8 | 8 | 0.4 |
| 4 | 6 | 6 | 6 | 6 | 0.4 |
| 5 | 3 | 7 | 3 | 7 | 0.4 |
| 6 | 2 | 9 | 2 | 9 | 0.6 |
| 7 | 5 | 5 | 3 | 3 | 0.3 |
| 8 | 8 | 1 | 8 | 1 | 0.7 |
| 9 | 6 | 2 | 6 | 2 | 0.5 |
| 10 | 7 | 3.6 | 7 | 3.6 | 0.5 |

**Table 9. 3D Surface Plots of the Benchmark Functions F1-F23**

| F1 | F2 | F3 | F4 |
|---|---|---|---|
| 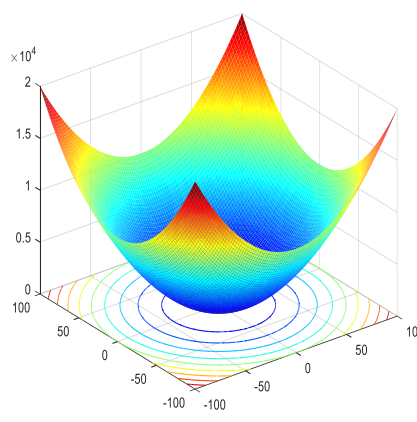 | 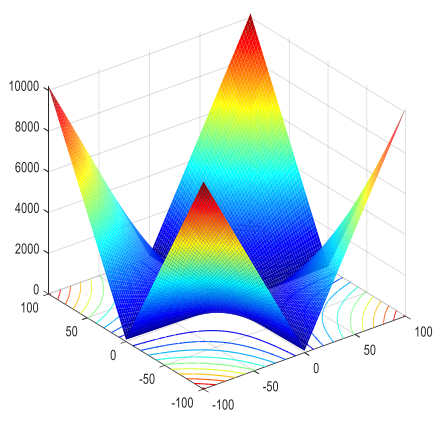 | 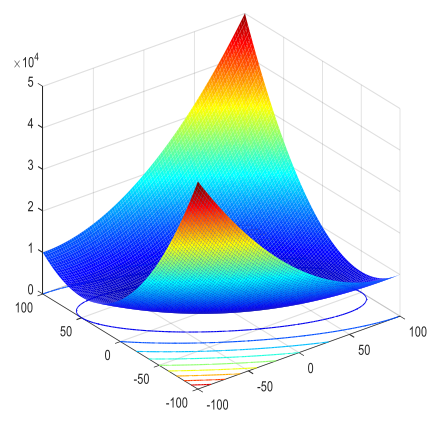 | 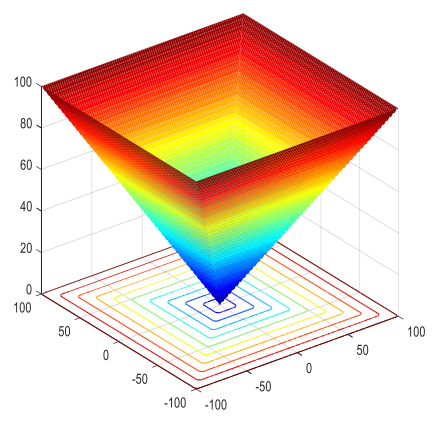 |
| **F5** | **F6** | **F7** | **F8** |
| 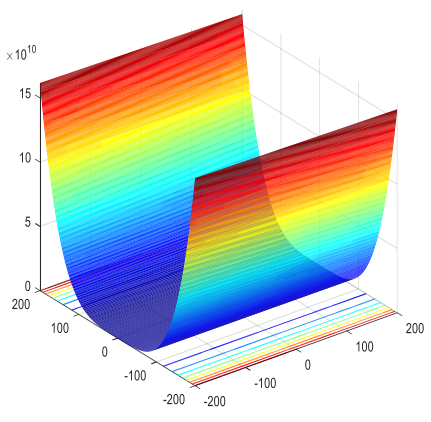 | 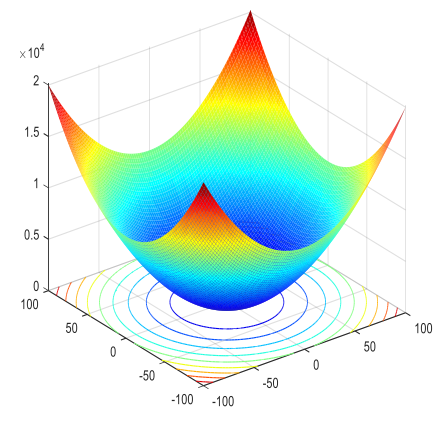 | 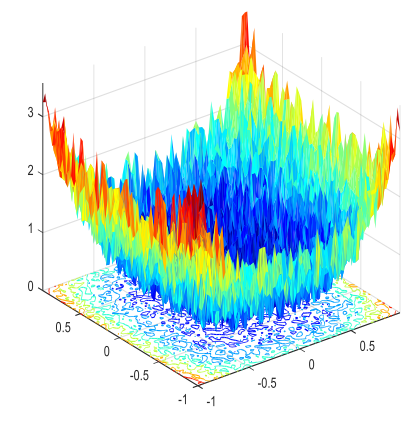 | 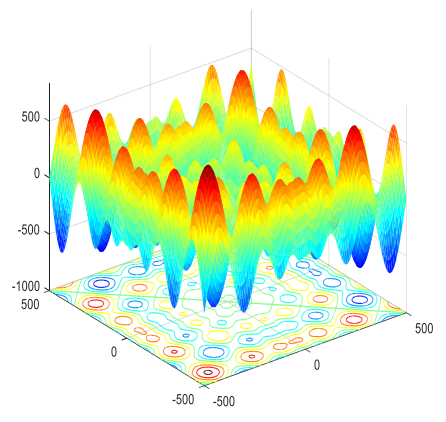 |
| **F9** | **F10** | **F11** | **F12** |
| 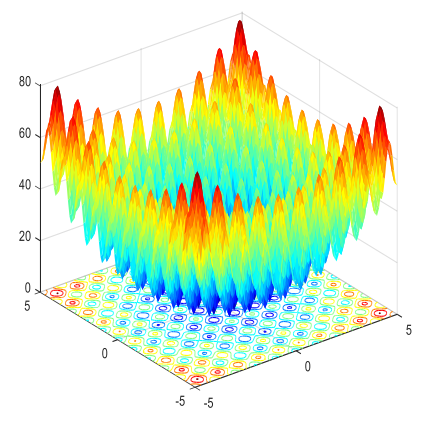 | 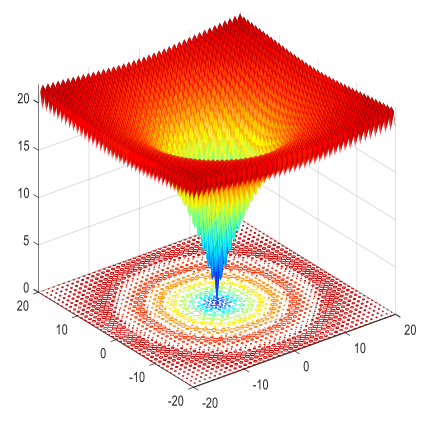 | 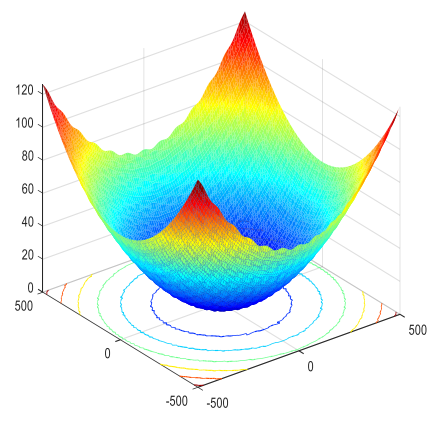 | 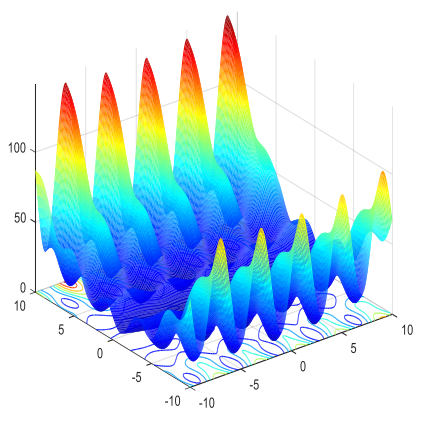 |
| **F13** | **F14** | **F15** | **F16** |
| 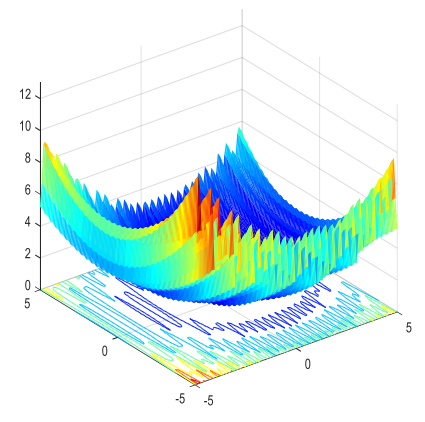 | 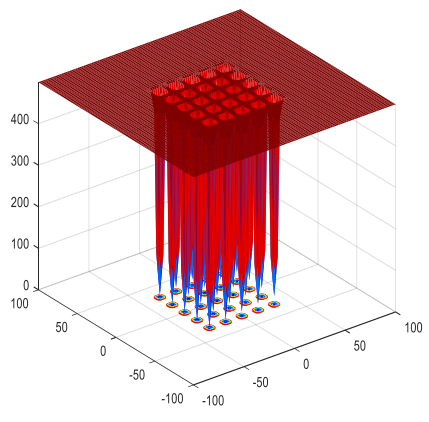 | 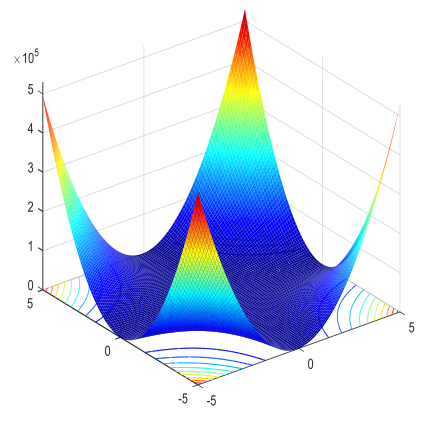 | 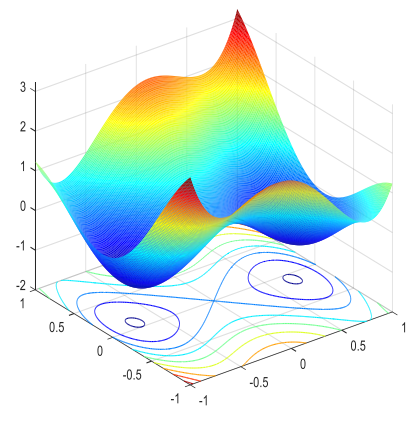 |
| **F17** | **F18** | **F19** | **F20** |
| 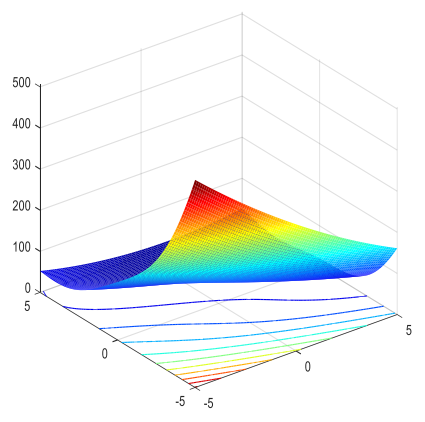 | 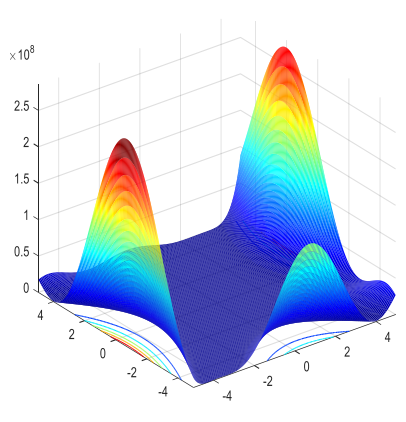 | 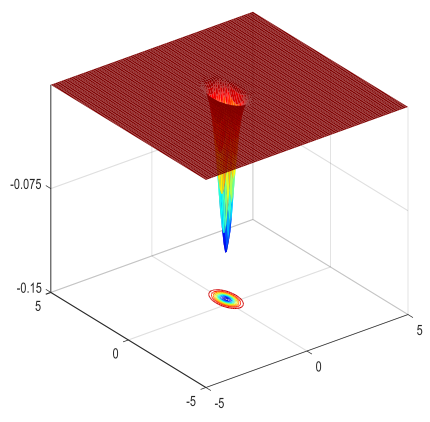 | 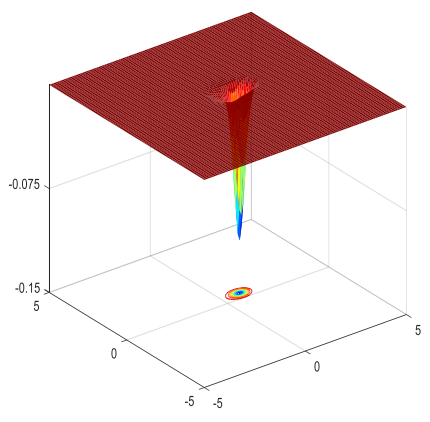 |
| **F21** | **F22** | **F23** | |
| 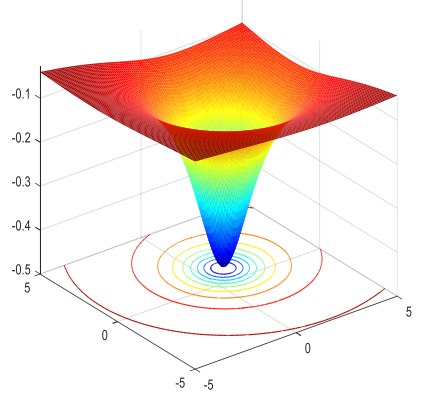 | 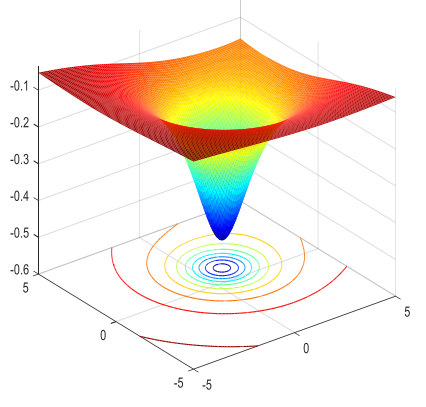 | 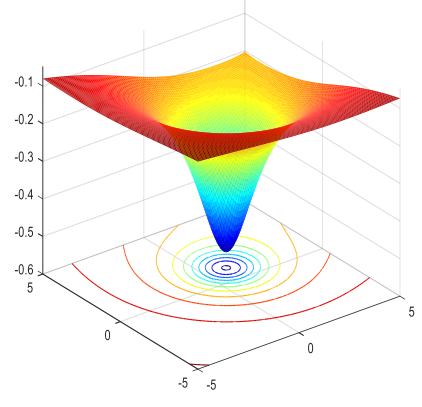 | |

*5.2. Parameter Settings*

We choose the constant $k$ to be 5 and θ to be the product of a random number drawn from a zero-mean Gaussian distribution with a standard deviation of 0.5 and a factor of the order of $10^{-3}$ after sufficient number of trials. The learning rate $\chi$ decreases linearly with iterations from 1 to 0.3 according to equation (24) as an LTV weight [8]. $\rho^{Chebyshev} \in [0,1]$ is a random number generated using a Chebyshev chaotic map in equation (27). All experiments are carried out on two Intel(R) Core(TM) i7-5500U CPUs @ 2.40GHz with 8GB RAM and one Intel(R) Core(TM) i7-2600U CPU @ 3.40GHz with 16GB RAM using MATLAB R2017a. All experiments are independently repeated 30 times in order to account for variability in reported data due to the underlying stochasticity of the metaheuristics used. Clusters from the MATLAB Parallel Computing Cloud are utilized to speed up the benchmarking.

## 6. Experimental Results

Tables 10 through 12 report performances of the C-QDDS algorithm on the test problems stacked against solution qualities obtained using 8 other commonly used, recent nature-inspired approaches. These are: i) Sine Cosine Algorithm (SCA) [18], ii) Dragon Fly Algorithm (DFA) [19], iii) Ant Lion Optimization (ALO) [20], iv) Whale Optimization Algorithm (WOA) [21], v) Firefly Algorithm (FA) [22], vi) Quantum-behaved Particle Swarm Optimization (QPSO) [2-3], vii) Particle Swarm Optimization with Damped Inertia * (PSO-I) and viii) the canonical Particle Swarm Optimizer (PSO-II) [6]. Each algorithm has been executed for 1000 iterations with 30 independent trials following which their mean, standard deviation and minimum values are noted. Testing carried out on the functions adhere to the dimensionalities and range constraints specified in Tables 2-8. A total of 50 agents have been introduced in the particle pool, out of which only one agent is picked in each iteration. The rationale for choosing one agent instead of many or all from the pool is to investigate the incremental effect of a single agent's propagation under different nature-inspired dynamical perturbations. The ripple effect caused otherwise, by many sensor reading exchanges among many or all particles may be delayed when a single particle affects the global pool of particles in one iteration.

* PSO-I utilizes an exponentially decaying inertia weight for exploration-exploitation trade-off

*6.1. Test Results on Optimization Problems*

**Table 10. Solution Quality in Unimodal Functions in Table 2 (30D, 1000 Iterations over 30 Independent Trials)**

| Fn | Stat | C-QDDS Chebyshev Map | Sine Cosine Algorithm | Dragon Fly Algorithm | Ant Lion Optimization | Whale Optimization | Firefly Algorithm | QPSO | PSO w=0.95*w | PSO No Damping |
|---|---|---|---|---|---|---|---|---|---|---|
| F1 | Mean | 1.1956e-06 | 0.0055 | 469.8818 | **7.8722e-07** | 17.3824 | 3.5794e+04 | 3.0365e+03 | 109.5486 | 110.3989 |
| | Min | 5.1834e-07 | 1.0207e-07 | 23.9914 | **8.9065e-08** | 0.6731 | 3.0236e+04 | 1.3286e+03 | 39.3329 | 42.8825 |
| | Std | **2.8711e-07** | 0.0161 | 474.0822 | 1.0286e-06 | 19.6687 | 3.3373e+03 | 920.4817 | 43.3127 | 54.7791 |
| F2 | Mean | 0.0051 | **3.6862-06** | 9.2230 | 27.8542 | 0.7846 | 3.4566e+04 | 36.4162 | 4.2299 | 4.4102 |
| | Min | 0.0025 | 2.7521e-09 | **0** | 0.0029 | 0.0745 | 84.8978 | 21.7082 | 2.0290 | 2.1627 |
| | Std | 9.7281e-04 | **8.9681e-06** | 5.7226 | 42.2856 | 0.5303 | 1.3595e+05 | 12.5312 | 1.1111 | 1.3804 |
| F3 | Mean | **1.0265e-04** | 3.4383e+03 | 6.3065e+03 | 302.3783 | 1.0734e+05 | 4.4017e+04 | 3.0781e+04 | 4.0409e+03 | 3.4218e+03 |
| | Min | **1.0184e-05** | 27.3442 | 310.7558 | 102.7732 | 5.0661e+04 | 3.0021e+04 | 1.8940e+04 | 2.2416e+03 | 1.9223e+03 |
| | Std | **6.5905e-05** | 3.1641e+03 | 4.7838e+03 | 167.7687 | 4.0661e+04 | 6.6498e+03 | 5.9848e+03 | 994.2550 | 997.4284 |
| F4 | Mean | **3.6945e-04** | 12.8867 | 13.8222 | 8.8157 | 66.4261 | 68.4102 | 56.5926 | 12.8272 | 11.9252 |
| | Min | **1.4162e-04** | 1.4477 | 4.1775 | 2.0212 | 17.8904 | 62.9296 | 32.6744 | 10.2302 | 9.0857 |
| | Std | **9.8034e-05** | 8.1625 | 5.5197 | 3.0808 | 21.5187 | 2.6497 | 8.2985 | 1.5793 | 2.0508 |
| F5 | Mean | **28.7211** | 60.7787 | 2.0123e+04 | 143.9657 | 1.5976e+03 | 7.4584e+07 | 2.1204e+06 | 6.0590e+03 | 5.3377e+03 |
| | Min | 28.7074 | 28.0932 | 44.0682 | **20.7989** | 39.9132 | 3.8917e+07 | 5.0759e+05 | 655.5618 | 1.2610e+03 |
| | Std | **0.0077** | 55.2793 | 3.6793e+04 | 288.1879 | 3.0458e+03 | 2.0606e+07 | 9.1390e+05 | 4.2558e+03 | 2.6303e+03 |
| F6 | Mean | 7.2332 | 4.2963 | 488.3942 | **6.0117e-07** | 30.0158 | 3.6216e+04 | 3.6028e+03 | 107.5196 | 116.9431 |
| | Min | 6.4389 | 3.3201 | 17.4978 | **8.9390e-08** | 0.8531 | 2.8838e+04 | 1.8380e+03 | 45.9374 | 28.9258 |
| | Std | 0.5612 | 0.4007 | 309.2795 | **6.2634e-07** | 44.1595 | 2.8434e+03 | 986.7972 | 47.5633 | 49.5767 |
| F7 | Mean | **0.0037** | 0.0289 | 0.1491 | 0.0541 | 0.1265 | 36.0335 | 1.4761 | 0.1737 | 0.1749 |
| | Min | **4.9685e-04** | 0.0010 | 0.0157 | 0.0210 | 0.0177 | 21.1334 | 0.3837 | 0.0697 | 0.0734 |
| | Std | **0.0023** | 0.0472 | 0.0918 | 0.0229 | 0.0993 | 7.5632 | 0.7718 | 0.0561 | 0.0690 |

**Table 11. Solution Quality in Multimodal Functions in Table 3 (30D, 1000 Iterations over 30 Independent Trials)**

| | | C-QDDS Chebyshev Map | Sine Cosine Algorithm | Dragon Fly Algorithm | Ant Lion Optimizer | Whale Optimization | Firefly Algorithm | QPSO | PSO w=0.95*w | PSO No Damping |
|---|---|---|---|---|---|---|---|---|---|---|
| F8 | Mean | -602.2041 | -4.0397e+03 | -6.001e+03 | -5.5942e+03 | **-8.5061e+03** | -3.8714e+03 | -3.3658e+03 | -5.1487e+03 | -4.8821e+03 |
| | Best | -975.5422 | -4.4739e+03 | -8.9104e+03 | -8.2843e+03 | **-1.0768e+04** | -4.2603e+03 | -5.0298e+03 | -7.4208e+03 | -6.6643e+03 |
| | Std | **160.8409** | 214.0523 | 783.7255 | 515.1599 | 895.4642 | 204.0029 | 486.0400 | 766.3330 | 750.3092 |
| F9 | Mean | **2.4873e-04** | 8.8907 | 124.0432 | 79.9945 | 116.4796 | 328.4011 | 248.0831 | 57.8114 | 57.1125 |
| | Best | 8.2194e-05 | **1.0581e-06** | 32.1699 | 45.7681 | 0.4305 | 308.3590 | 177.8681 | 19.1318 | 27.4985 |
| | Std | **6.3770e-05** | 16.2284 | 40.4730 | 22.2932 | 88.0344 | 10.1050 | 31.9501 | 15.1644 | 15.0292 |
| F10 | Mean | **8.1297e-04** | 10.7873 | 6.0693 | 1.6480 | 1.1419 | 19.3393 | 12.3433 | 4.9951 | 4.9271 |
| | Best | 5.6777e-04 | 3.4267e-05 | **8.8818e-16** | 1.7296e-04 | 0.0265 | 18.4515 | 9.7835 | 3.9874 | 2.9208 |
| | Std | **8.8526e-05** | 9.6938 | 1.9141 | 0.9544 | 0.9926 | 0.2797 | 1.8413 | 0.6230 | 0.7957 |
| F11 | Mean | **8.7473e-08** | 0.1770 | 5.0784 | 0.0082 | 1.1735 | 316.5026 | 33.5446 | 2.0669 | 2.0604 |
| | Best | **3.5705e-08** | 2.2966e-05 | 1.1727 | 2.5498e-05 | 0.9839 | 226.5205 | 11.9701 | 1.3636 | 1.3744 |
| | Std | **2.6504e-08** | 0.2195 | 4.5098 | 0.0093 | 0.2340 | 33.3806 | 12.5605 | 0.5989 | 0.5366 |
| F12 | Mean | **0.0995** | 991.4301 | 12.2571 | 9.4380 | 642.0404 | 1.2629e+08 | 5.6147e+05 | 6.4329 | 6.5610 |
| | Best | **0** | 0.2878 | 1.6755 | 3.4007 | 0.0442 | 5.6104e+07 | 4.0841e+04 | 1.0266 | 2.8742 |
| | Std | **0.2621** | 5.4201e+03 | 13.5218 | 3.9121 | 3.5039e+03 | 4.4034e+07 | 6.9761e+05 | 2.7882 | 3.0003 |
| F13 | Mean | **0.0105** | 3.1940 | 1.5156e+04 | 0.0133 | 2.3405e+03 | 2.8867e+08 | 3.5568e+06 | 38.1945 | 39.0369 |
| | Best | **0** | 1.8776 | 5.6609 | 2.7212e-07 | 0.3813 | 1.3101e+08 | 6.8216e+05 | 12.7653 | 15.4619 |
| | Std | 0.0576 | 2.2922 | 6.0811e+04 | **0.0163** | 1.2373e+04 | 8.1766e+07 | 2.4393e+06 | 15.2922 | 27.6751 |

**Table 12. Solution Quality in Multimodal Functions in Table 4 (Fixed Dim, 1000 Iters over 30 Independent Trials)**

| | | C-QDDS Chebyshev Map | Sine Cosine Algorithm | Dragon Fly Algorithm | Ant Lion Optimizer | Whale Optimization | Firefly Algorithm | QPSO | PSO w=0.95*w | PSO No Damping |
|---|---|---|---|---|---|---|---|---|---|---|
| F14, n=2 | Mean | 3.6771 | 1.3949 | **1.0311** | 1.2299 | 4.2524 | 1.0519 | 2.3561 | 2.7786 | 3.7082 |
| | Best | 1.0056 | **0.9980** | **0.9980** | **0.9980** | **0.9980** | **0.9980** | 0.9981 | **0.9980** | **0.9980** |
| | Std | 2.2295 | 0.8072 | **0.1815** | 0.4276 | 3.7335 | 0.1889 | 1.7188 | 2.2246 | 2.7536 |
| F15, n=4 | Mean | **3.7361e-04** | 9.1075e-04 | 0.0016 | 0.0027 | 0.0051 | 0.0024 | 0.0030 | 0.0036 | 0.0034 |
| | Best | 3.1068e-04 | 3.1549e-04 | 4.7829e-04 | 4.0518e-04 | 3.4820e-04 | 0.0011 | 7.2169e-04 | 3.6642e-04 | **3.0858e-04** |
| | Std | **5.0123e-05** | 4.2242e-04 | 0.0014 | 0.0060 | 0.0076 | 0.0012 | 0.0059 | 0.0063 | 0.0068 |
| F16, n=2 | Mean | -0.5487 | **-1.0316** | **-1.0316** | **-1.0316** | -1.0315 | -1.0295 | **-1.0316** | **-1.0316** | **-1.0316** |
| | Best | -1.0315 | **-1.0316** | **-1.0316** | **-1.0316** | **-1.0316** | **-1.0316** | **-1.0316** | **-1.0316** | **-1.0316** |
| | Std | 0.4275 | 1.1863e-05 | 1.4229e-06 | **3.6950e-14** | 3.3613e-04 | 0.0030 | 1.1009e-04 | 8.2108e-14 | 2.7251e-13 |
| F17, n=2 | Mean | 0.4721 | 0.3983 | **0.3979** | **0.3979** | 0.4069 | 0.4002 | 0.4000 | **0.3979** | **0.3979** |
| | Best | 0.3989 | **0.3979** | **0.3979** | **0.3979** | **0.3979** | **0.3979** | **0.3979** | **0.3979** | **0.3979** |
| | Std | 0.0920 | 4.8435e-04 | 4.9327e-08 | **2.3588e-14** | 0.0179 | 0.0020 | 0.0043 | 5.0770e-10 | 2.1067e-08 |
| F18, n=2 | Mean | 3.8438 | **3** | **3** | **3** | 3.9278 | 3.0402 | 3.0007 | **3.0000** | **3.0000** |
| | Best | 3.0080 | **3** | **3** | **3** | **3.0000** | 3.0002 | **3.0000** | **3.0000** | **3.0000** |
| | Std | 0.9128 | 5.7657e-06 | 8.7817e-07 | **1.2869e-13** | 5.0752 | 0.0397 | 0.0017 | 1.0155e-11 | 5.8511e-11 |
| F19, n=3 | Mean | -3.6805 | -3.8547 | -3.8625 | **-3.8628** | -3.8246 | -3.8542 | **-3.8628** | **-3.8628** | **-3.8628** |
| | Best | -3.8587 | -3.8626 | **-3.8628** | **-3.8628** | **-3.8628** | -3.8625 | **-3.8628** | **-3.8628** | **-3.8628** |
| | Std | 0.1942 | 0.0016 | 8.8455e-04 | **7.5193e-15** | 0.0657 | 0.0066 | 1.5043e-05 | 5.2841e-11 | 9.2140e-11 |
| F20, n=6 | Mean | -2.2207 | -2.9961 | -3.2421 | **-3.2705** | -3.0966 | -3.0645 | -3.2646 | -3.2625 | -3.2546 |
| | Best | -2.7562 | -3.2911 | **-3.3220** | **-3.3220** | -3.2610 | -3.2436 | -3.3219 | **-3.3220** | **-3.3220** |
| | Std | 0.29884 | 0.2060 | 0.0670 | **0.0599** | 0.1535 | 0.0911 | 0.0605 | 0.0605 | **0.0599** |
| F21, n=4 | Mean | -3.1126 | -4.0962 | **-9.0360** | -6.7752 | -6.5291 | -4.3198 | -5.8537 | -5.3955 | -5.4045 |
| | Best | -4.5610 | -5.3343 | **-10.1532** | **-10.1532** | -9.8465 | -7.5958 | -10.1474 | **-10.1532** | **-10.1532** |
| | Std | **0.7090** | 1.5519 | 1.9130 | 2.6824 | 1.9988 | 1.4599 | 3.5651 | 3.3029 | 3.4897 |
| F22, n=4 | Mean | -3.2009 | -3.9949 | **-10.0455** | -7.2979 | -6.3611 | -4.2776 | -6.7830 | -5.3236 | -6.3098 |
| | Best | -4.5933 | -7.9241 | **-10.4029** | **-10.4029** | -10.2432 | -9.2741 | -10.3974 | **-10.4029** | **-10.4029** |
| | Std | **0.7098** | 2.1774 | 1.3422 | 3.0440 | 2.3852 | 1.6527 | 3.5783 | 3.2000 | 3.4602 |
| F23, n=4 | Mean | -2.3595 | -4.6650 | **-9.9928** | -7.1691 | -5.2592 | -4.6959 | -7.5372 | -7.3175 | -5.1501 |
| | Best | -4.2043 | -7.7259 | **-10.5364** | **-10.5364** | -10.0617 | -8.5734 | -10.5344 | **-10.5364** | **-10.5364** |
| | Std | **0.8183** | 1.5038 | 1.6439 | 3.2926 | 2.5389 | 1.4647 | 3.6778 | 3.7753 | 3.4033 |

**Table 13. Win/Tie/Loss Count among Competitors w.r.t Reported Global Best**

| Performance | Metric | C-QDDS Chebyshev Map | Sine Cosine Algorithm | Dragon Fly Algorithm | Ant Lion Optimizer | Whale Optimization | Firefly Algorithm | QPSO | PSO w=0.95*w | PSO No Damping |
|---|---|---|---|---|---|---|---|---|---|---|
| Win | Mean | **10** | 1 | 3 | 3 | 1 | 0 | 0 | 0 | 0 |
| | Best | **6** | 1 | 2 | 3 | 1 | 0 | 0 | 0 | 1 |
| | Std | **14** | 1 | 1 | 7 | 0 | 0 | 0 | 0 | 0 |
| Tie | Mean | 0 | 2 | 3 | **4** | 0 | 0 | 2 | **4** | 5 |
| | Best | 0 | 4 | **9** | **9** | 5 | 3 | 4 | **9** | **9** |
| | Std | **0** | **0** | **0** | **0** | **0** | **0** | **0** | **0** | **0** |
| Lose | Mean | **13** | 20 | 17 | 17 | 22 | 23 | 21 | 19 | 18 |
| | Best | 17 | 18 | **12** | 13 | 18 | 20 | 19 | 14 | 13 |
| | Std | **9** | 22 | 22 | 17 | 23 | 23 | 23 | 23 | 23 |

**Table 14. Average Ranks based on Win/Tie/Loss Count among Competitors w.r.t Reported Global Best**

| Performance | Metric | C-QDDS Chebyshev Map | Sine Cosine Algorithm | Dragon Fly Algorithm | Ant Lion Optimizer | Whale Optimization | Firefly Algorithm | QPSO | PSO w=0.95*w | PSO No Damping |
|---|---|---|---|---|---|---|---|---|---|---|
| Win | Mean | **1** | 3 | 2 | 2 | 3 | 4 | 4 | 4 | 4 |
| | Best | **1** | 4 | 3 | 2 | 4 | 5 | 5 | 5 | 4 |
| | Std | **1** | 3 | 3 | 2 | 4 | 4 | 4 | 4 | 4 |
| Tie | Mean | 5 | 4 | 3 | **2** | 5 | 5 | 4 | **2** | **1** |
| | Best | 5 | 3 | 2 | 2 | 3 | 4 | 3 | 2 | **1** |
| | Std | 1 | 1 | 1 | **1** | 1 | 1 | 1 | 1 | **1** |
| Lose | Mean | **1** | 5 | 2 | 2 | 7 | 8 | 6 | 4 | 3 |
| | Best | 4 | 5 | **1** | **2** | 5 | 7 | 6 | 3 | **2** |
| | Std | **1** | 3 | 3 | 2 | 4 | 4 | 4 | 4 | 3 |
| Average Rank | Mean | **2.333** | 4 | **2.333** | 2 | 5 | 5.666 | 4.666 | 3.333 | 2.666 |
| | Best | 3.333 | 4 | **2** | **2** | 4 | 5.333 | 4.666 | 3.333 | 2.333 |
| | Std | **1** | 2.333 | 2.333 | 1.666 | 3 | 3 | 3 | 3 | 2.666 |

**Table 15. Results of Two-tailed t-test for C-QDDS v/s Competitors**

| **Algorithm** | **C-QDDS vs SCA** | **C-QDDS vs DFA** | **C-QDDS vs ALO** | **C-QDDS vs WOA** | **C-QDDS vs FA** | **C-QDDS vs QPSO** | **C-QDDS vs PSO-II** | **C-QDDS vs PSO-I** |
|---|---|---|---|---|---|---|---|---|
| **Function** | **t values ($t_{critical}$ = 2.001717). Null Hypothesis: ($\mu_{CQDDS} - \mu_{Competitor}$) > 0** | | | | | | | |
| **F1** | -1.8707 | -5.4287 | 2.094532 | -4.84055 | -58.7456 | -18.0684 | -13.8533 | -11.0385 |
| **F2** | 28.69263 | -8.82265 | -3.60728 | -8.05108 | -1.39261 | -15.9148 | -20.8264 | -17.4788 |
| **F3** | -5.95188 | -7.22065 | -9.87189 | -14.4592 | -36.2554 | -28.1704 | -22.2608 | -18.7903 |
| **F4** | -8.64702 | -13.7155 | -15.6724 | -16.9076 | -141.411 | -37.3523 | -44.4852 | -31.8485 |
| **F5** | -3.17636 | -2.99135 | -2.19031 | -2.8213 | -19.825 | -12.7079 | -7.76098 | -11.0552 |
| **F6** | 23.32769 | -8.52117 | 70.59491 | -2.82556 | -69.7487 | -19.9572 | -11.5478 | -12.12 |
| **F7** | -2.92082 | -8.67254 | -11.9943 | -6.77163 | -26.0926 | -10.4491 | -16.5837 | -13.5823 |
| **F8** | 70.32003 | 36.96027 | 50.66345 | 47.58374 | 68.92735 | 29.56635 | 31.80233 | 30.54904 |
| **F9** | -3.0006 | -16.7868 | -19.6538 | -7.24698 | -178.004 | -42.529 | -20.8808 | -20.8139 |
| **F10** | -6.09462 | -17.3651 | -9.45308 | -6.29659 | -378.696 | -36.7146 | -43.9082 | -33.9102 |
| **F11** | -4.41671 | -6.1678 | -4.82933 | -27.4681 | -51.933 | -14.6277 | -18.9028 | -21.0311 |
| **F12** | -1.00178 | -4.92371 | -13.0453 | -1.00347 | -15.7087 | -4.40833 | -12.3869 | -11.7511 |
| **F13** | -7.60459 | -1.36509 | -0.25619 | -1.03608 | -19.337 | -7.98647 | -13.6763 | -7.72376 |
| **F14** | 5.271808 | 6.47901 | 5.904436 | -0.72462 | 6.426319 | 2.570191 | 1.562548 | -0.04808 |
| **F15** | -6.9162 | -4.79494 | -2.12362 | -3.40618 | -9.2411 | -2.4381 | -2.80494 | -2.43761 |
| **F16** | 6.187023 | 6.187023 | 6.187023 | 6.18574 | 6.159965 | 6.187023 | 6.187023 | 6.187023 |
| **F17** | 4.393627 | 4.417501 | 4.417501 | 3.810236 | 4.27956 | 4.287797 | 4.417501 | 4.417501 |
| **F18** | 5.063193 | 5.063193 | 5.063193 | -0.08922 | 4.81742 | 5.058984 | 5.063193 | 5.063193 |
| **F19** | 4.912978 | 5.133083 | 5.141597 | 3.849854 | 4.896216 | 5.141597 | 5.141597 | 5.141597 |
| **F20** | 11.70106 | 18.26704 | 18.86578 | 14.28008 | 14.7933 | 18.75247 | 18.71474 | 18.58005 |
| **F21** | 3.157566 | 15.90258 | 7.230404 | 8.823441 | 4.074111 | 4.13039 | 3.701433 | 3.525209 |
| **F22** | 1.898948 | 24.69127 | 7.179345 | 6.955444 | 3.278706 | 5.378252 | 3.547072 | 4.820763 |
| **F23** | 7.375911 | 22.76815 | 7.76455 | 5.953976 | 7.627314 | 7.526917 | 7.029854 | 4.366702 |
| **Significantly better** | **9** | **12** | **10** | **11** | **12** | **13** | **13** | **13** |
| **Significantly worse** | **11** | **10** | **12** | **8** | **10** | **10** | **9** | **9** |

## Table 16: Cohen's d values for C-QDDS v/s Competitors

| **Algorithm** | **C-QDDS vs SCA** | **C-QDDS vs DFA** | **C-QDDS vs ALO** | **C-QDDS vs WOA** | **C-QDDS vs FA** | **C-QDDS vs QPSO** | **C-QDDS vs PSO-II** | **C-QDDS vs PSO-I** |
|---|---|---|---|---|---|---|---|---|
| **Function** | **Cohen's d values, where** $d = \frac{\mu_{C-QDDS} - \mu_{Competitor}}{\sqrt{\frac{s^2_{\mu_CQDDS} + s^2_{\mu_Competitor}}{2}}}$ | | | | | | | |
| F1 | -0.483 | -1.4017 | 0.5408 | -1.2498 | -15.1681 | -4.6652 | -3.5769 | -2.8501 |
| F2 | 7.4084 | -2.278 | -0.9314 | -2.0788 | -0.3596 | -4.1092 | -5.3773 | -4.513 |
| F3 | -1.5368 | -1.8644 | -2.5489 | -3.7333 | -9.3611 | -7.2736 | -5.7477 | -4.8516 |
| F4 | -2.2327 | -3.5413 | -4.0466 | -4.3655 | -36.5121 | -9.6443 | -11.486 | -8.2233 |
| F5 | -0.8201 | -0.7724 | -0.5655 | -0.7285 | -5.1188 | -3.2812 | -2.0039 | -2.8544 |
| F6 | 6.0232 | -2.2002 | 18.2275 | -0.7296 | -18.009 | -5.1529 | -2.9816 | -3.1294 |
| F7 | -0.7542 | -2.2392 | -3.0969 | -1.7484 | -6.7371 | -2.698 | -4.2819 | -3.5069 |
| F8 | 18.1566 | 9.5431 | 13.0812 | 12.2861 | 17.797 | 7.634 | 8.2113 | 7.8877 |
| F9 | -0.7748 | -4.3343 | -5.0746 | -1.8712 | -45.9603 | -10.9809 | -5.3914 | -5.3741 |
| F10 | -1.5736 | -4.4836 | -2.4408 | -1.6258 | -97.7789 | -9.4797 | -11.3371 | -8.7556 |
| F11 | -1.1404 | -1.5925 | -1.2469 | -7.0922 | -13.4091 | -3.7769 | -4.8807 | -5.4302 |
| F12 | -0.2587 | -1.2713 | -3.3683 | -0.2591 | -4.056 | -1.1382 | -3.1983 | -3.0341 |
| F13 | -1.9635 | -0.3525 | -0.0661 | -0.2675 | -4.9928 | -2.0621 | -3.5312 | -1.9943 |
| F14 | 1.3612 | 1.6729 | 1.5245 | -0.1871 | 1.6593 | 0.6636 | 0.4034 | -0.0124 |
| F15 | -1.7858 | -1.238 | -0.5483 | -0.8795 | -2.386 | -0.6295 | -0.7242 | -0.6294 |
| F16 | 1.5975 | 1.5975 | 1.5975 | 1.5972 | 1.5905 | 1.5975 | 1.5975 | 1.5975 |
| F17 | 1.1344 | 1.1406 | 1.1406 | 0.9838 | 1.105 | 1.1071 | 1.1406 | 1.1406 |
| F18 | 1.3073 | 1.3073 | 1.3073 | -0.023 | 1.2439 | 1.3062 | 1.3073 | 1.3073 |
| F19 | 1.2685 | 1.3254 | 1.3276 | 0.994 | 1.2642 | 1.3276 | 1.3276 | 1.3276 |
| F20 | 3.0212 | 4.7165 | 4.8711 | 3.6871 | 3.8196 | 4.8419 | 4.8321 | 4.7973 |
| F21 | 0.8153 | 4.106 | 1.8669 | 2.2782 | 1.0519 | 1.0665 | 0.9557 | 0.9102 |
| F22 | 0.4903 | 6.3753 | 1.8537 | 1.7959 | 0.8466 | 1.3887 | 0.9158 | 1.2447 |
| F23 | 1.9045 | 5.8787 | 2.0048 | 1.5373 | 1.9694 | 1.9434 | 1.8151 | 1.1275 |

## Table 17: Hedges' g values for C-QDDS v/s Competitors

| Algorithm | C-QDDS vs SCA | C-QDDS vs DFA | C-QDDS vs ALO | C-QDDS vs WOA | C-QDDS vs FA | C-QDDS vs QPSO | C-QDDS vs PSO-II | C-QDDS vs PSO-I |
|---|---|---|---|---|---|---|---|---|
| **Function** | **Hedge's g values, where** $g = \frac{\mu_{C-QDDS} - \mu_{Competitor}}{\sqrt{\frac{(n_1-1) * s^2_{\mu_CQDDS} + (n_2-1) * s^2_{\mu_Competitor}}{n_1+n_2-2}}}$ | | | | | | | |
| F1 | -0.6716 | -1.949 | 0.752 | -1.7378 | -21.0904 | -6.4867 | -4.9735 | -3.9629 |
| F2 | 10.301 | -3.1674 | -1.2951 | -2.8905 | -0.5 | -5.7136 | -7.4768 | -6.2751 |
| F3 | -2.1368 | -2.5923 | -3.5441 | -5.1909 | -13.0161 | -10.1135 | -7.9919 | -6.7459 |
| F4 | -3.1044 | -4.924 | -5.6266 | -6.07 | -50.768 | -13.4099 | -15.9706 | -11.434 |
| F5 | -1.1403 | -1.074 | -0.7863 | -1.0129 | -7.1174 | -4.5623 | -2.7863 | -3.9689 |
| F6 | 8.3749 | -3.0593 | 25.3443 | -1.0145 | -25.0405 | -7.1648 | -4.1457 | -4.3513 |
| F7 | -1.0487 | -3.1135 | -4.3061 | -2.4311 | -9.3676 | -3.7514 | -5.9537 | -4.8761 |
| F8 | 25.2457 | 13.2691 | 18.1887 | 17.0831 | 24.7457 | 10.6146 | 11.4173 | 10.9674 |
| F9 | -1.0773 | -6.0266 | -7.0559 | -2.6018 | -63.9052 | -15.2683 | -7.4964 | -7.4724 |
| F10 | -2.188 | -6.2342 | -3.3938 | -2.2606 | -135.956 | -13.181 | -15.7636 | -12.1742 |
| F11 | -1.5857 | -2.2143 | -1.7337 | -9.8613 | -18.6446 | -5.2516 | -6.7863 | -7.5504 |
| F12 | -0.3597 | -1.7677 | -4.6834 | -0.3603 | -5.6396 | -1.5826 | -4.4471 | -4.2187 |
| F13 | -2.7301 | -0.4901 | -0.0919 | -0.3719 | -6.9422 | -2.8672 | -4.9099 | -2.773 |
| F14 | 1.8927 | 2.3261 | 2.1197 | -0.2602 | 2.3072 | 0.9227 | 0.5609 | -0.0172 |
| F15 | -2.4831 | -1.7214 | -0.7624 | -1.2229 | -3.3176 | -0.8753 | -1.007 | -0.8751 |
| F16 | 2.2212 | 2.2212 | 2.2212 | 2.2208 | 2.2115 | 2.2212 | 2.2212 | 2.2212 |
| F17 | 1.5773 | 1.5859 | 1.5859 | 1.3679 | 1.5364 | 1.5394 | 1.5859 | 1.5859 |
| F18 | 1.8177 | 1.8177 | 1.8177 | -0.032 | 1.7296 | 1.8162 | 1.8177 | 1.8177 |
| F19 | 1.7638 | 1.8429 | 1.846 | 1.3821 | 1.7578 | 1.846 | 1.846 | 1.846 |
| F20 | 4.2008 | 6.558 | 6.773 | 5.1267 | 5.3109 | 6.7324 | 6.7188 | 6.6704 |
| F21 | 1.1336 | 5.7092 | 2.5958 | 3.1677 | 1.4626 | 1.4829 | 1.3288 | 1.2656 |
| F22 | 0.6817 | 8.8645 | 2.5775 | 2.4971 | 1.1771 | 1.9309 | 1.2734 | 1.7307 |
| F23 | 2.6481 | 8.174 | 2.7876 | 2.1375 | 2.7383 | 2.7022 | 2.5238 | 1.5677 |

## Table 18: Precision Plots (Fraction of Successful Runs v/s Cost Range) for the 23 Benchmark Functions

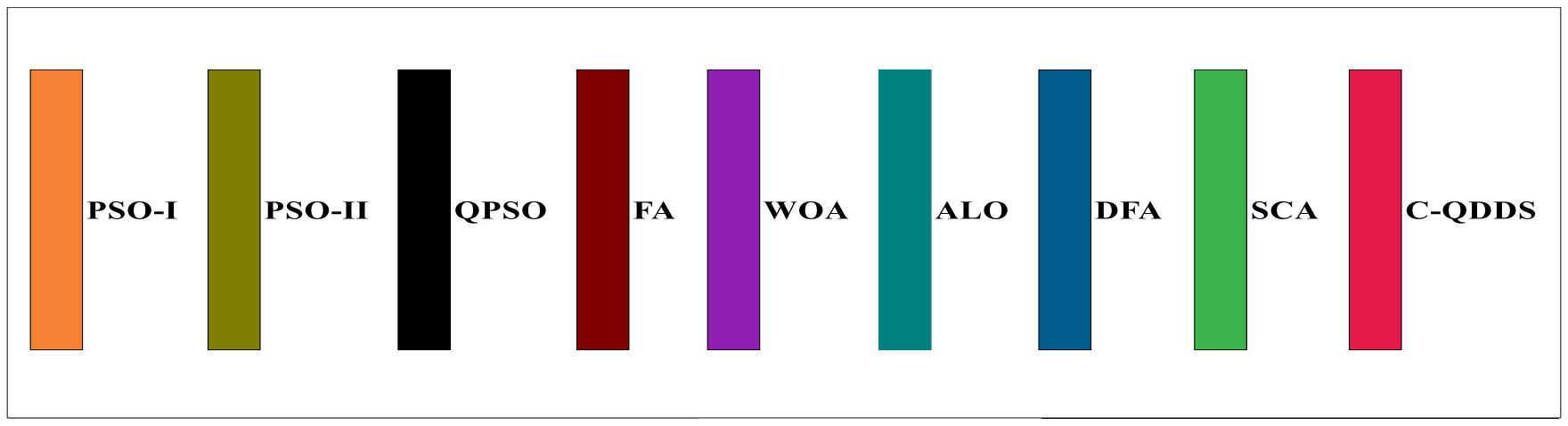

**F1**

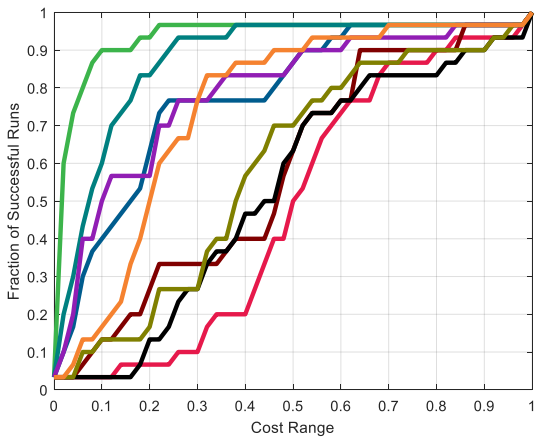

**F2**

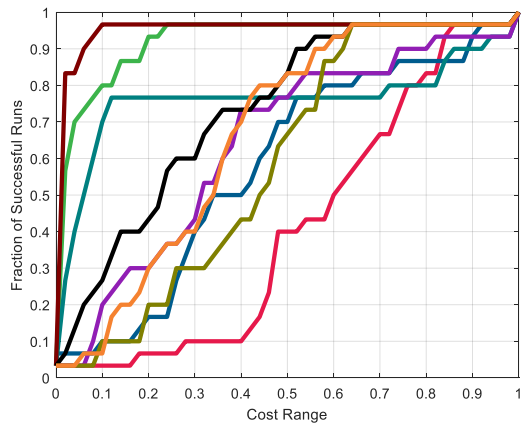

**F3**

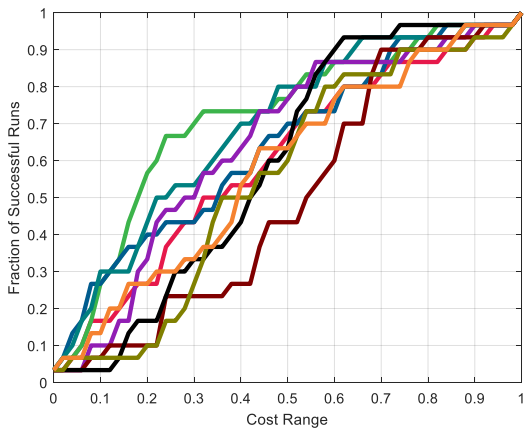

**F4**

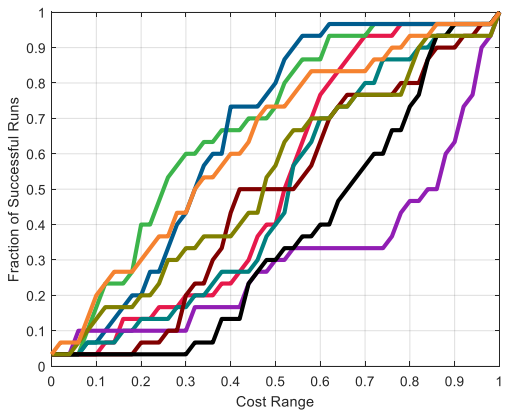

**F5**

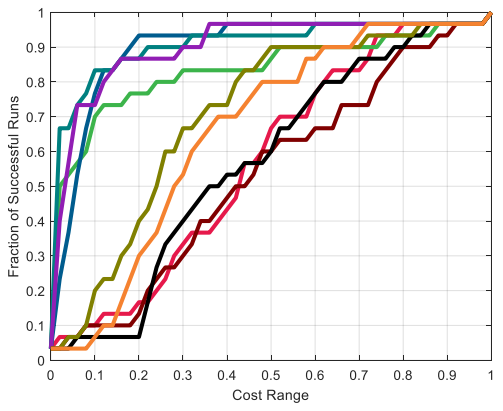

**F6**

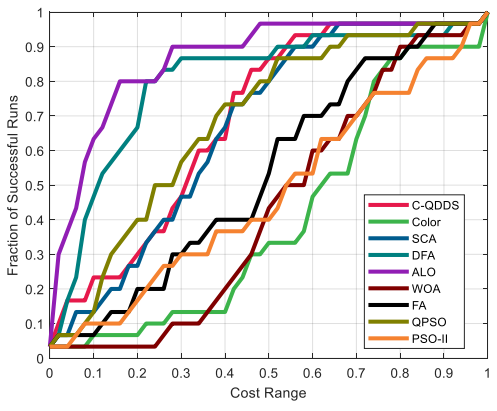

**F7**

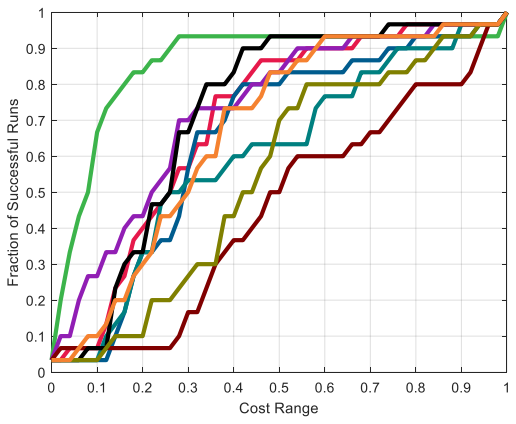

**F8**

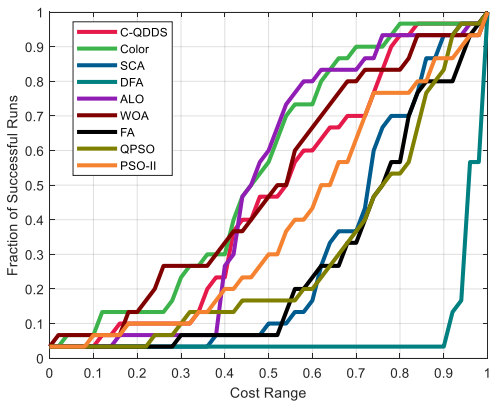

**F9**

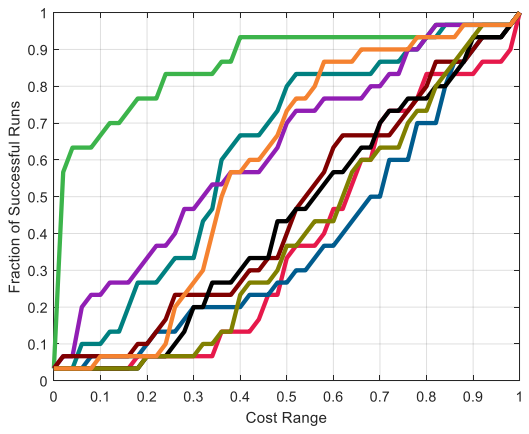

**F10**

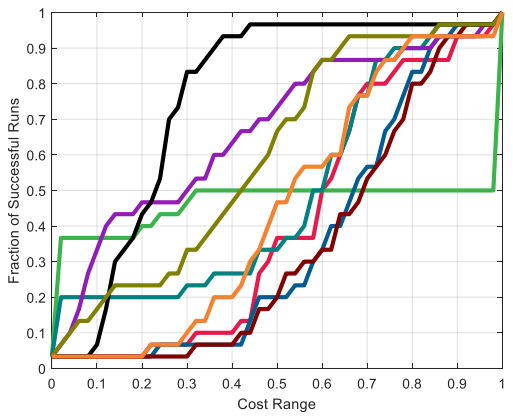

**F11**

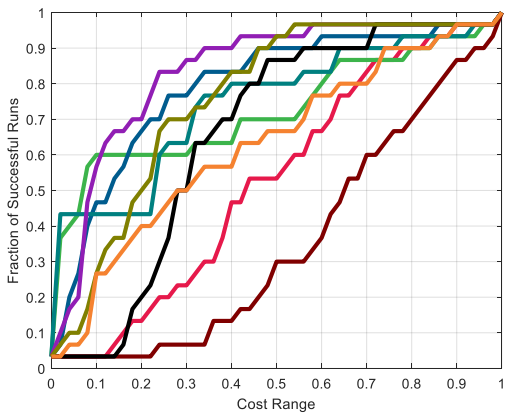

**F12**

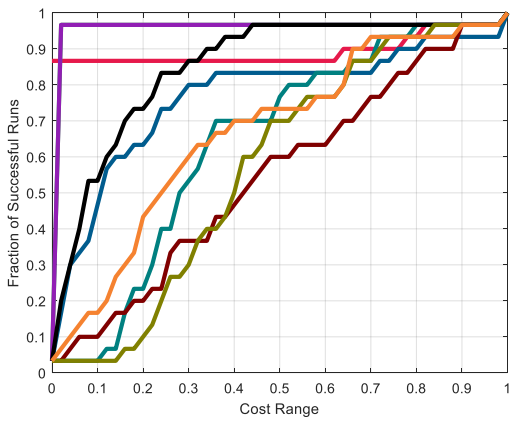

**F13**

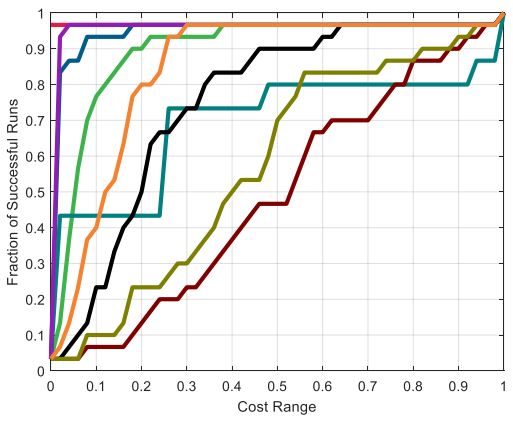

**F14**

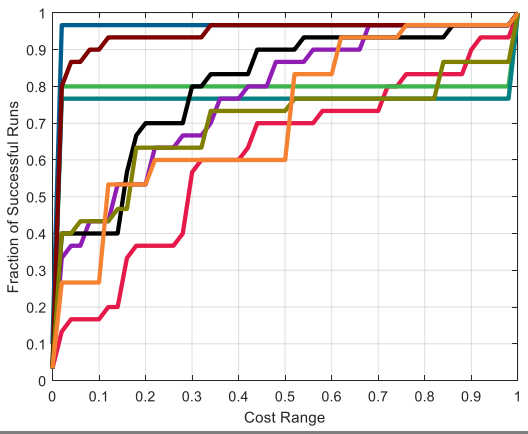

**F15**

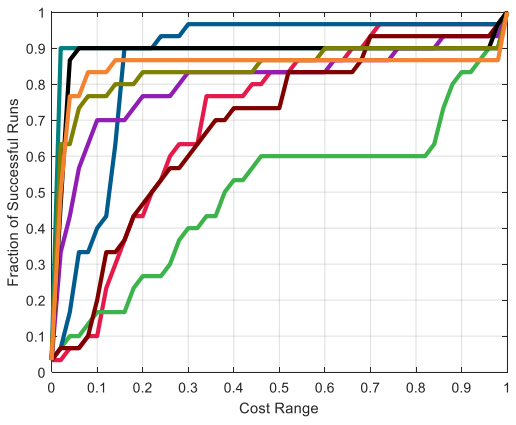

**F16**

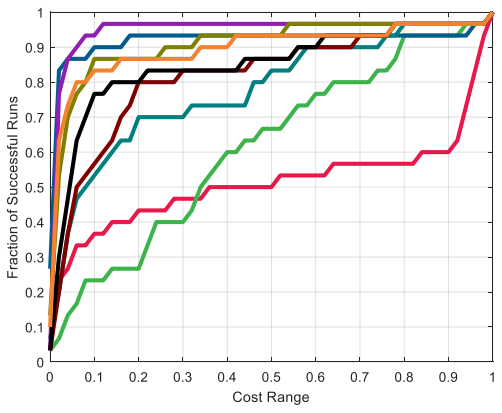

**F17**

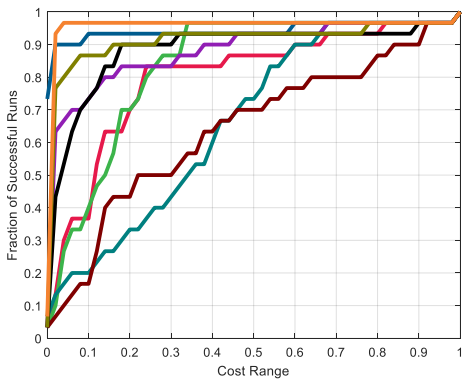

**F18**

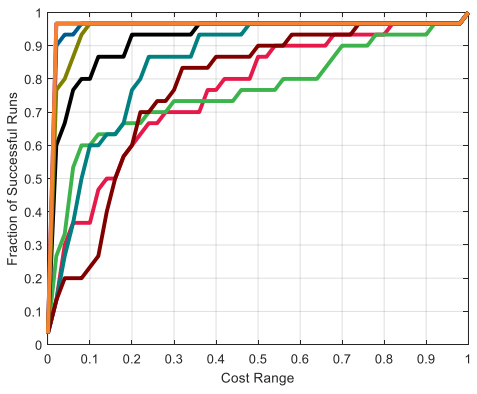

**F19**

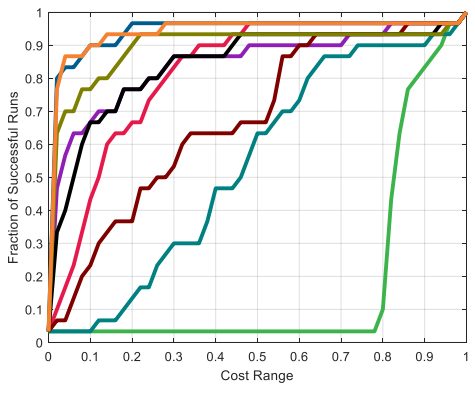

**F20**

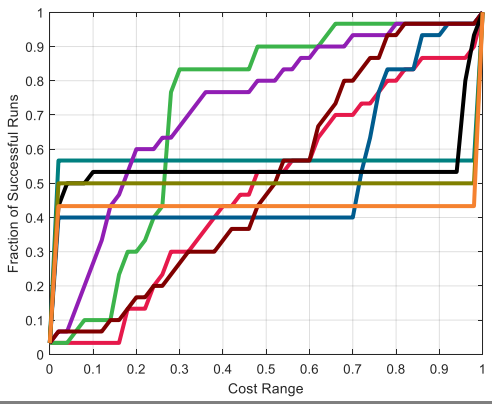

**F21**

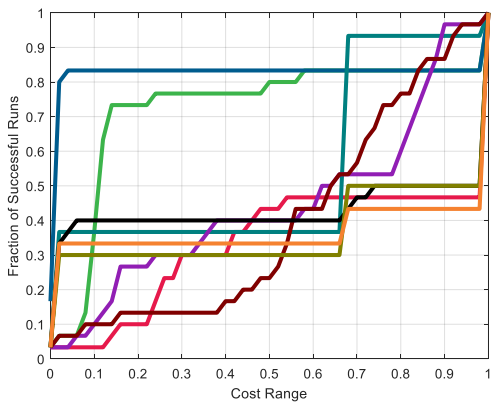

**F22**

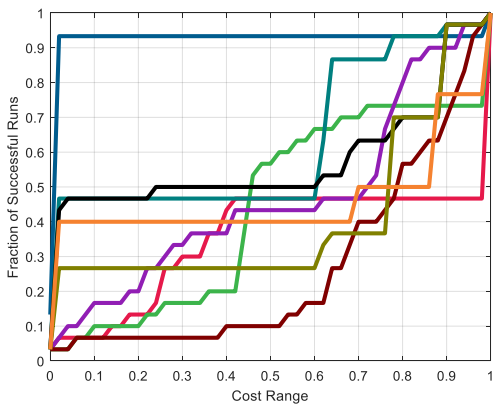

**F23**

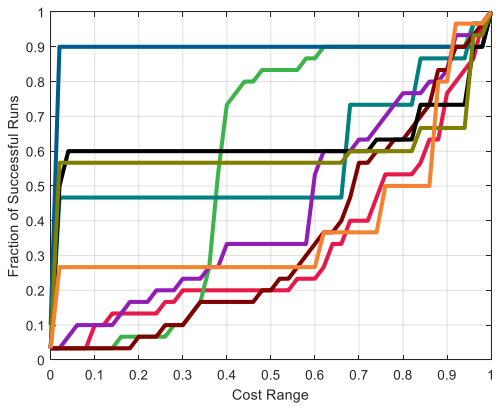

**Table 19: Trajectory of the Best Solutions for the 23 Benchmark Functions**

**F1**

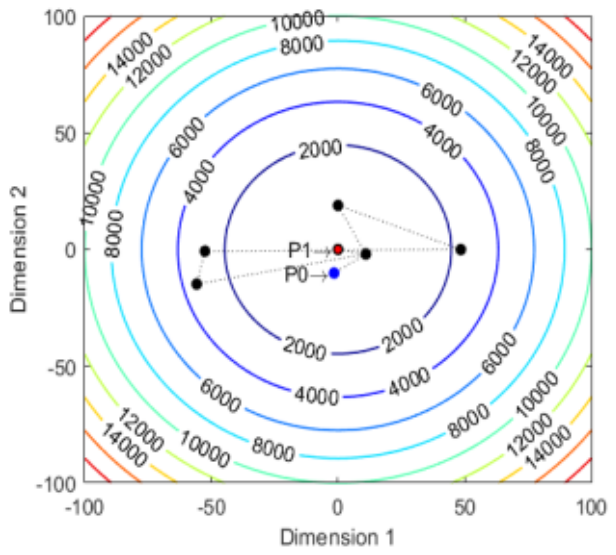

**F2**

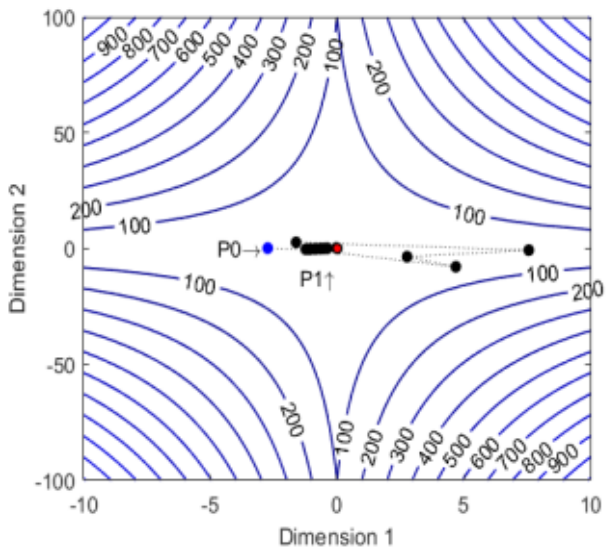

**F3**

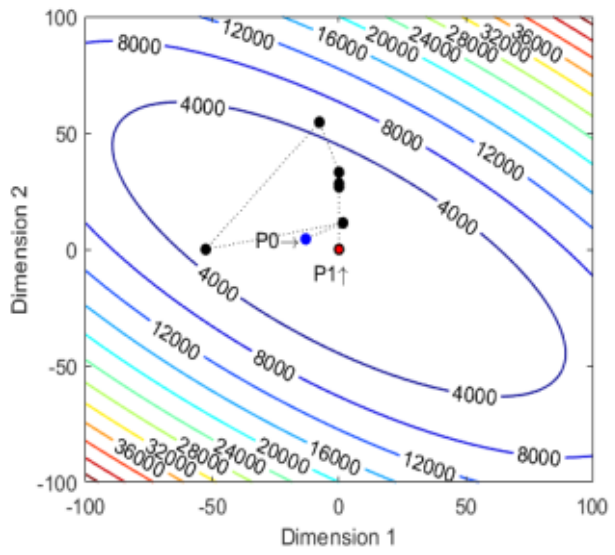

**F4**

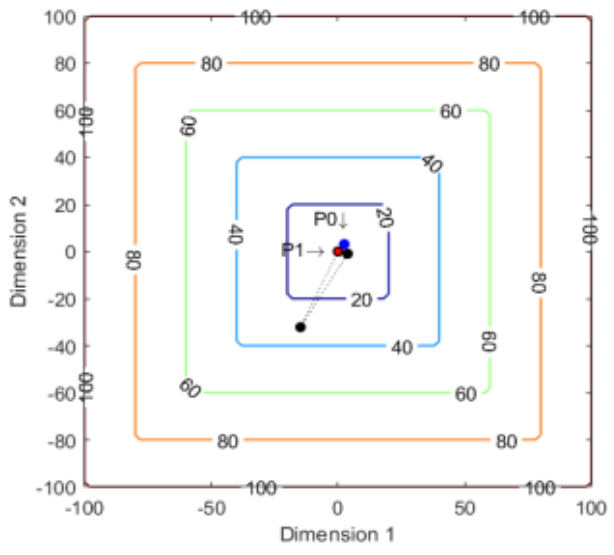

**F5**

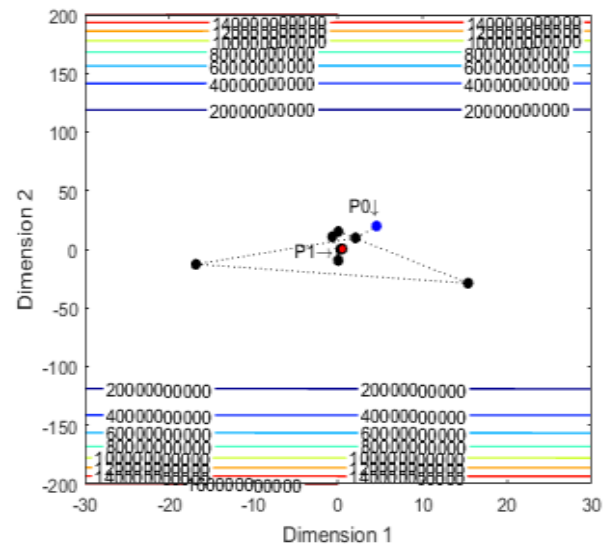

**F6**

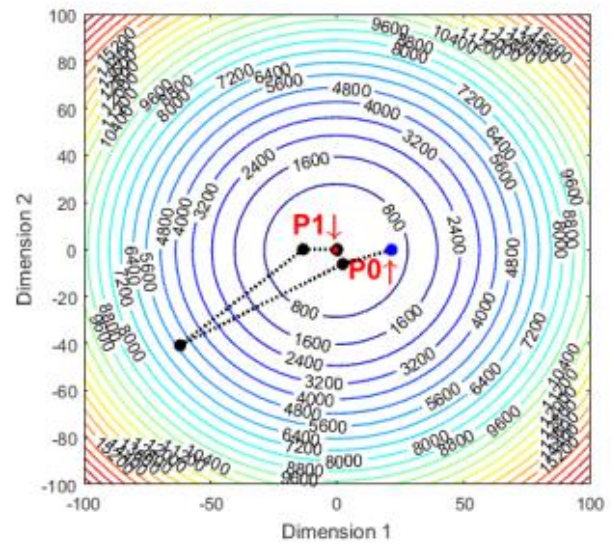

**F7**

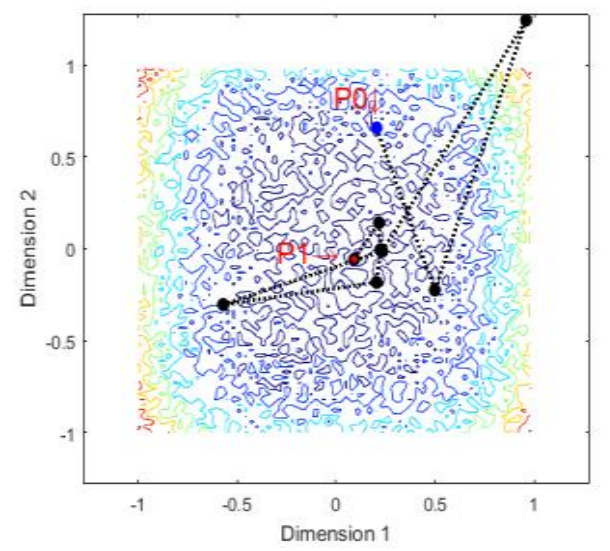

**F8**

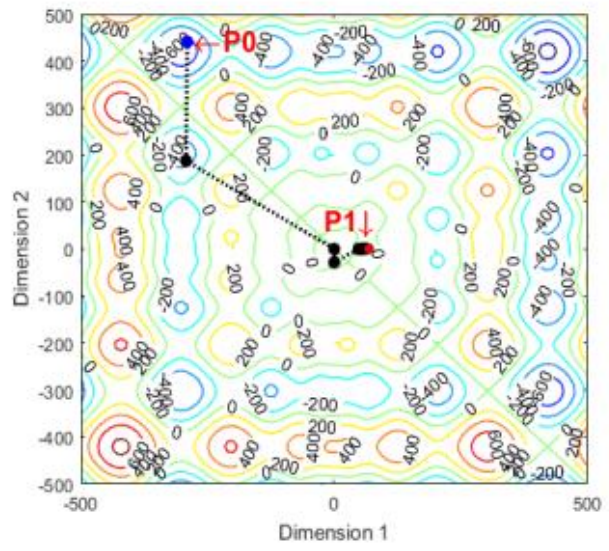

**F9**

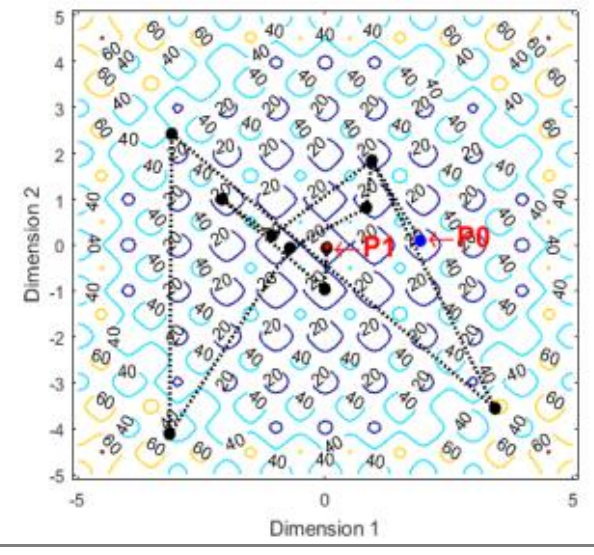

**F10**

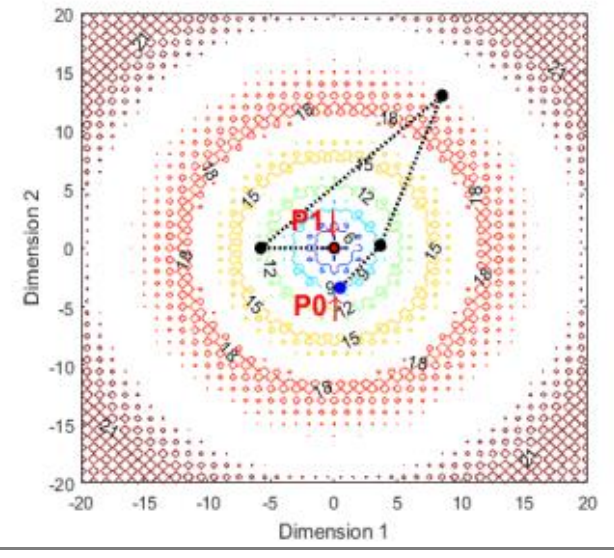

**F11**

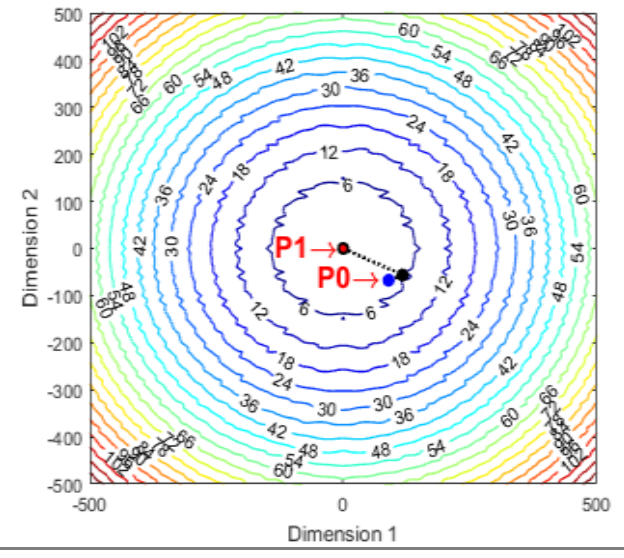

**F12**

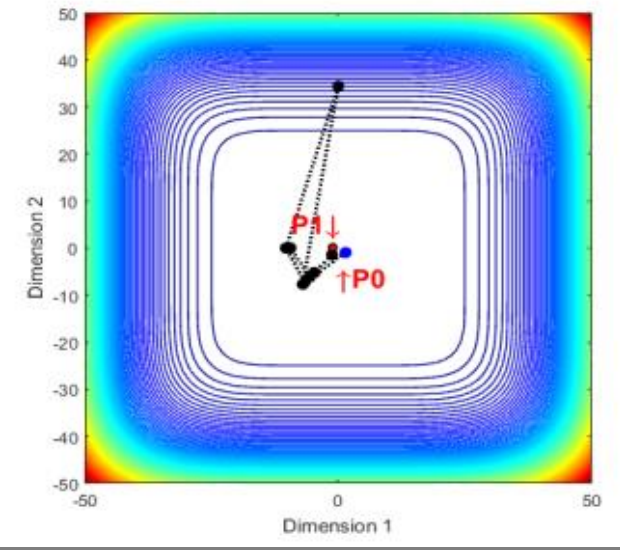

**F13**

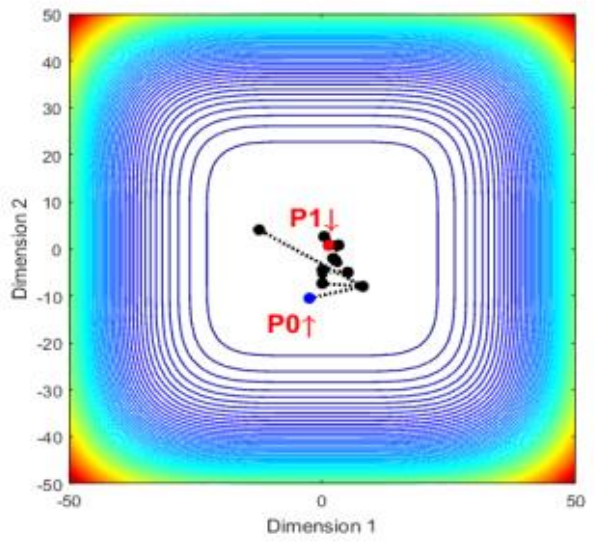

**F14**

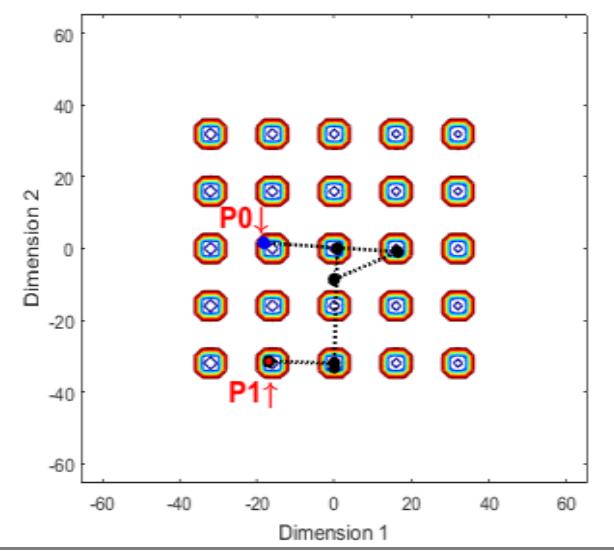

**F15**

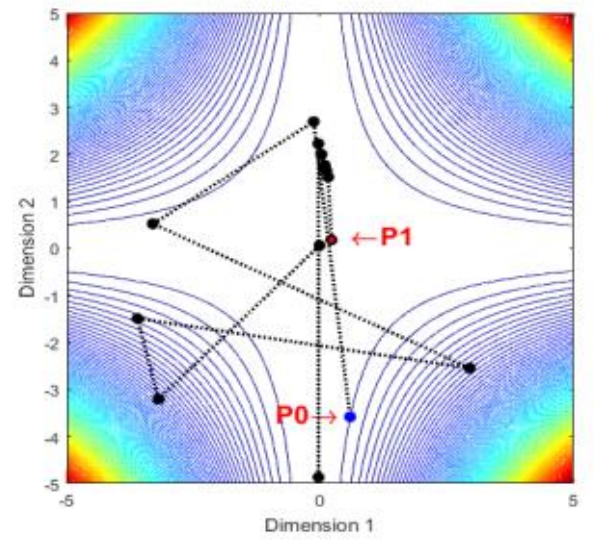

**F16**

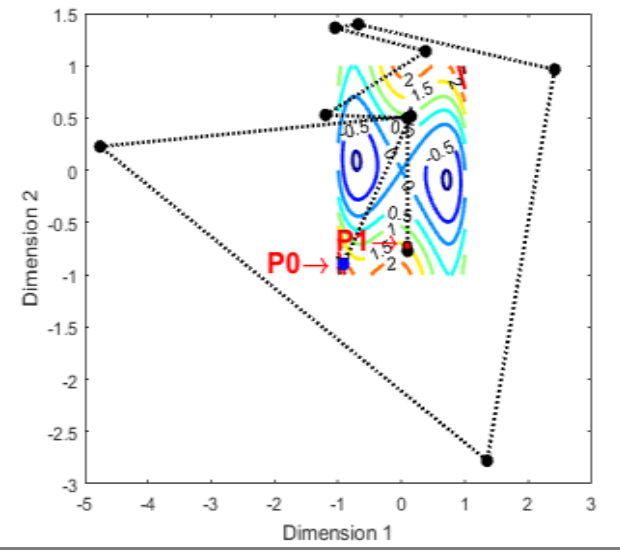

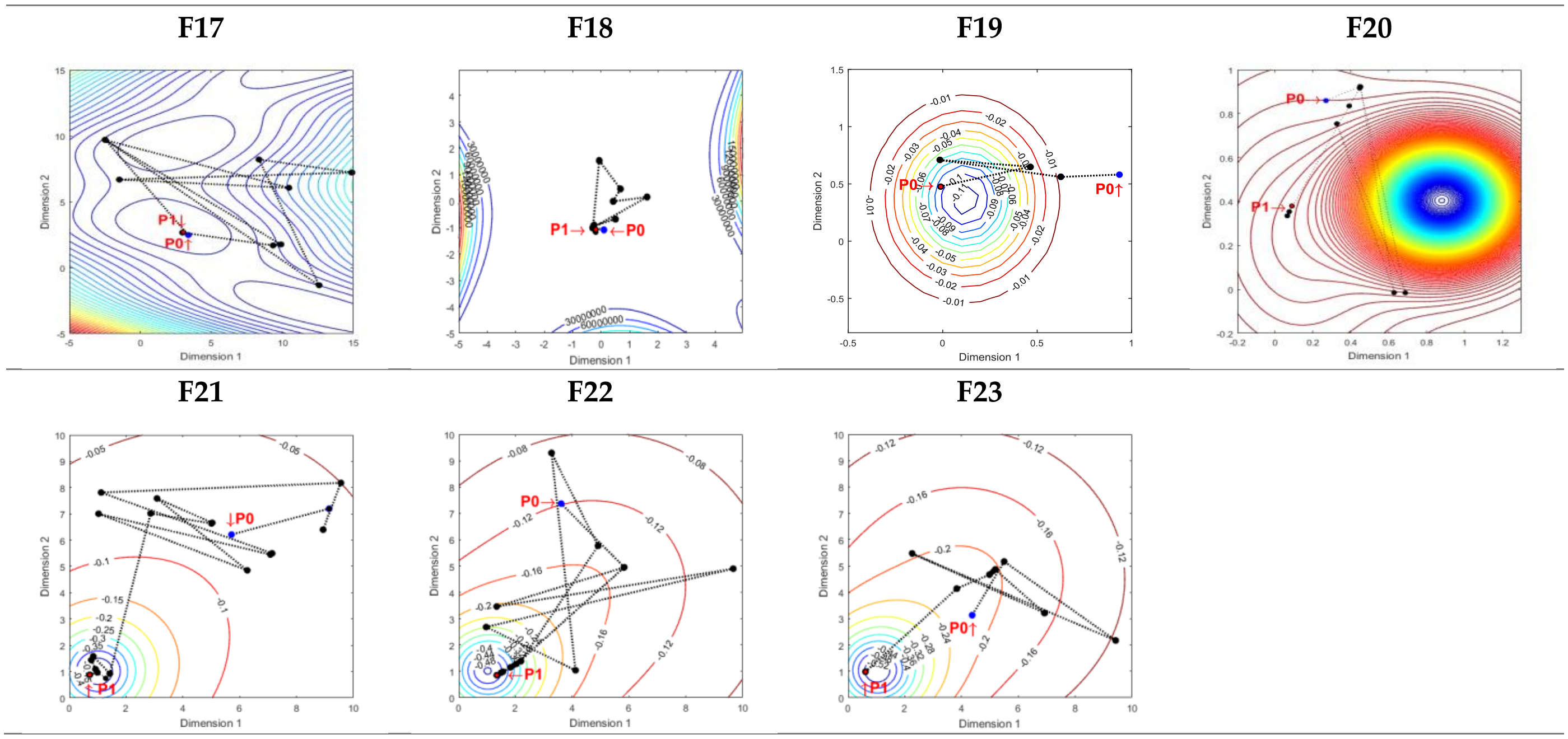

## 7. Analysis of Experimental Results

Tables 10-12 report the solution qualities obtained on the suite of test functions F1-F23 followed by Tables 13-17 in which the win/tie/loss counts, average ranks and results of statistical significance tests such as that of a two-tailed t-test and Cohen's d and Hedge's g values are reported. From tables 10-12 one can make the observation that C-QDDS has a distinctive advantage over the other algorithms in terms of quality of optima found, outperforming competitors in unimodal functions as F3-F5, F7 and multimodal ones such as F9-F13. However, solution quality drops for multimodal functions F14-F23, with the agents getting stuck in local minima. One interpretation is that since communication between particles is limited when only one agent is drawn in an iteration, it will take a considerably large number of iterations for promising regions to be found. Alternatively, because the QDDS mechanism is based on gradient descent, saddle points and valleys introduce stagnation which is difficult to break out of. A two-tailed student's t-test with significance level $\Theta = 0.05$ in Table 15 is used to accept or reject the hypothesis that the performance of the C-QDDS algorithm is significant when compared to any of the other approaches. It is observed that in general, C-QDDS provides superior solution quality when applied to problems in Tables 10-11 and that the difference is statistically significant at $\Theta = 0.05$. A measure of the effect sizes is provided in Table 16 through the computation of Cohen's d values, however to account for the correction Hedge's g values have also been reported in Table 17.

In Table 18, the number of successful executions against the obtained cost range for any algorithm is demonstrated for all test functions. The horizontal axis represents a value equivalent to the sum of the lowest cost obtained during the 30 runs of an algorithm and a fraction of the cost range (i.e. highest cost – lowest cost) ranging from 0.1 through 1 at intervals of 0.1. The vertical axis is the cumulative number of trials that resulted in solutions with lower cost than the corresponding horizontal axis value. For example, the vertical axis value at the horizontal tick of 0.1 is the number of trials having cost values less than {minimum cost + 0.1 × (maximum cost – minimum cost)}. These curves are a measure of the variability of the algorithmic solutions within their reported cost ranges and an indicator of how top-heavy or bottom-heavy they are. It is important to note that the cost range for each algorithm is different on every test function execution and as such the curves are merely meant for an intuitive understanding of the variability of the solutions and not intended to provide any basis for comparison among the algorithms. Algorithms having the least standard deviation among the cohort are expected to have a uniform density of solutions in the cost range and as such should follow a roughly linear relationship between the variables in the horizontal and vertical axes. It may be noted that C-QDDS, which roughly follows this relationship indeed has the least standard deviation in many cases, specifically for 14 of the 23 functions as illustrated in Table 13. This is in congruence with the convergence profiles of QDDS in Figures 1 through 12 of [1] which point out that QDDS is fairly consistent in its ability to converge to local optima of acceptable quality in certain problems.

Table 19 shows the trajectory evolution of the global best position across the functional iterations for each test case using C-QDDS. For ease of visualization, the contours of the 30-dimensional functions as well as the obtained *gbest* i.e. global best solutions are plotted using only the first 2 dimensions. $P_0$ represents the initial *gbest* position and $P_1$ represents the *gbest* position upon convergence, given the convergence criteria. The interim *gbest* position transitions are shown by dotted lines. The solutions to the 23 test problems outlined in the paper are local minima, however the quality of solutions that the C-QDDS and QDDS algorithm provide to some of these problems are markedly better than those reported in some studies in the literature [4,23,24,25]. A logical next-step to improve the optima seeking capability of the QDDS/C-QDDS approach is to introduce a problem-independent random walk in the $\boldsymbol{\delta}$ recomputing step of the algorithm instead of using gradient descent.

**8. Notes on Convergence of the Algorithm**

In this section, we discuss the convergence characteristics of the QDDS algorithm by formulating the algorithmic objective as an optimization problem and proving hypotheses adherence under certain weak assumptions. We start by considering the following problem $\mathbb{C}$ :

$\mathbb{C}$: *Provided there is a function* $\boldsymbol{f}$ *from* $\mathbb{R}^{\boldsymbol{n}}$ *to* $\mathbb{R}$ *and that* $\boldsymbol{S}$ *is a subset of* $\mathbb{R}^{\boldsymbol{n}}$*, a solution* $\boldsymbol{x}$ *in* $\boldsymbol{S}$ *is sought such that* $\boldsymbol{x}$ *minimizes* $\boldsymbol{f}$ *on* $\boldsymbol{S}$ *or finds an acceptable approximation of the minimum of* $\boldsymbol{f}$ *on* $\boldsymbol{S}$*.*

A conditioned approach to solving $\mathbb{C}$ was proposed by Solis and Wet [5] which we describe below. The rest of the proof follows logically from [5] as has also been shown by Van den Bergh in [26] and Sun et al in [27].

| **Algorithm 2. A conditioned approach to solving $\mathbb{C}$ [5]** |
|---|
| **1:** Initialize $x^0$ in $S$ and set $e = 0$ |
| **2:** Generate $\xi^e$ from the sample space $(\mathbb{R}^{\mathbf{n}}, \mathbb{B}, \mathbb{T}_{\boldsymbol{e}})$ |
| **3:** Update $x^{e+1} = £(x^e, \xi^e)$, choose $\mathbb{T}_{\boldsymbol{e+1}}$, set $e = e+1$ and repeat Step 1. |

The mapping £ is the optimization algorithm and should satisfy the following two hypotheses $\mathbb{H}^1$ and $\mathbb{H}^2$ in order to theoretically be globally convergent.

**Hypothesis** $\mathbb{H}^1$:

$$f(£\,(x, \xi)) \le f(x) \text{ and if } \xi \in S \text{ then } f(£\,(x, \xi)) \le f(\xi) \tag{29}$$

The sequence $f(x_e)_{e=1}^{\infty}$ generated by £ must monotonically reach a stable value, i.e. the infimum, for the mapping to be a globally convergent one.

**Hypothesis** $\mathbb{H}^2$: For any Borel subset $A$ of $S$ with $\vartheta(A) > 0$, it can be proved that:

$$\prod_{k=0}^{\infty}\{1 - \mathbb{T}_e(A)\} = 0 \tag{30}$$

This means that if there exists a subset $A$ of $S$ with positive volume then the chance that upon generating random samples $\xi^e$ it will repeatedly miss $A$ is zero. Guided random search methods are conditioned, which implies $\mathbb{T}_e$ depends on $x^0, x^1, \ldots, x^{e-1}$ generated in the preceding iterations. Therefore, $\mathbb{T}_e(A)$ is a conditional probability measure.

**Definition** $\mathbb{D}^1$**:** Values close to the essential infimum $\sigma$ is generated by a set of points having a non-zero $\vartheta$ measure.

$$\sigma = \inf\{t: \vartheta[x \in S \mid f(x) < t] > 0\} \tag{31}$$

**Definition** $\mathbb{D}^2$**:** The acceptable solution range $\mathfrak{R}_{\varepsilon,\mathfrak{S}}$ for $\mathbb{P}$ is constructed around the essential infimum $\sigma$ with step size $\varepsilon$ and bounded support $\mathfrak{S}$.

$$\mathfrak{R}_{\varepsilon,\mathfrak{S}} = \begin{cases} x \in S \mid f(x) < \sigma + \varepsilon, & \sigma \in (-\infty, \infty) \\ x \in S \mid f(x) < \mathfrak{S}, & \sigma \text{ is infinite} \end{cases} \tag{32}$$

**Theorem** $\mathbb{T}^1$**:** The Global Convergence Theorem for Random Search Algorithms states that when $\mathbb{H}^1$ and $\mathbb{H}^2$ are satisfied on a measurable subset of $\mathbb{R}^{\boldsymbol{n}}$ for a measurable function $f$, the probability that the conditioned sequence $\{x_e\}_{e=1}^{\infty}$ generated by the algorithm lies within the acceptable solution range $\mathfrak{R}_{\varepsilon,\mathfrak{S}}$ for $\mathbb{P}$ is one.

$$lim_{e\to\infty} P(x_e \in \mathfrak{R}_\varepsilon) = 1 \tag{33}$$

*8.1. Notes on Theoretical Convergence of the QDDS algorithm*

Proposition $\mathbb{P}^1$**:** The QDDS algorithm satisfies hypothesis $\mathbb{H}^1$.

Let us consider the solution update stage of the QDDS algorithm. If a new solution is generated such that its fitness is better than the ones recorded so far (global best - *gbest*), it replaces the best solution and is stored in memory.

$$x_{i,e+1} = \pounds\ (x_{i,e}) \tag{34}$$

$$update\ (gbest, x_e) = \begin{cases} gbest, & fit(new) < fit(best) \\ x_e, & otherwise \end{cases} \tag{35}$$

This implies sequence $\{fit(gbest_e)\}_{e=1}^{\infty}$ is monotonically decreasing and $fit(\pounds\ (x_e, gbest_e)) \le fit(x_e)$. So $\mathbb{H}^1$ is satisfied.

Proposition $\mathbb{P}^2$**:** The QDDS algorithm satisfies hypothesis $\mathbb{H}^2$.

Recall that in equation (14) the even solutions to the double delta potential well setup take on the form given below:

$$\psi_{even}(r) = \begin{cases} \eta_2(1 + e^{2ka})e^{-kr} & r > a \\ \eta_2(e^{-kr} + e^{kr}) & -a < r < a \\ \eta_2(1 + e^{2ka})e^{kr} & r < -a \end{cases} \tag{36}$$

$$\lambda(r_{i,j,t}) = \psi^2{}_{even,i,j,t}(r) = \begin{cases} {\eta_2}^2\left(e^{-2kr} + e^{2k(2a-r)} + 2e^{4k(a-r)}\right) & r > a \\ {\eta_2}^2(e^{-2kr} + e^{2kr} + 2) & -a < r < a \\ {\eta_2}^2\left(e^{2kr} + e^{2k(2a+r)} + 2e^{4k(a+r)}\right) & r < -a \end{cases} \tag{37}$$

$\psi^2{}_{even,i,j,t}(r)$ is a measure of the probability density function of a particle in a particular dimension and integrating it across all dimensions yields the corresponding cumulative distribution function $\Lambda_{i,t}(Set)$:

$$\Lambda_{i,t}(Set) = \int_{Set} \left\{\prod_{j=1}^{d} \lambda\left(r_{i,j,t}\right)\right\} dr_{i,1,t}dr_{i,2,t} \dots dr_{i,D,t} \tag{38}$$

Observe that when $r \to \pm\infty$, the probability measure $\psi^2{}_{even}(r)$ goes to zero for $r \in (-\infty, -a) \cup (a, \infty)$ and is bounded for the region $-a < r < a$.

$$\lim_{r\to\pm\infty} \lambda(r_{i,j,t}) = 0 \tag{39}$$

$$\therefore 0 < \Lambda(Set) < 1 \tag{40}$$

$$\Lambda_t(Set) = \cup_{i=1}^{n} \Lambda_{i,t} \tag{41}$$

$$\therefore \prod_{t=0}^{\infty}\{1 - \Lambda_t(Set)\} = 0 \tag{42}$$

Thus, $\mathbb{H}^2$ is also satisfied. This in turn implies that theorem $\mathbb{T}^1$, which is the global convergence algorithm for random search algorithms is also satisfied and £ is globally convergent.

**9. Concluding Remarks**

The Chaotic Quantum Double Delta Swarm (C-QDDS) Algorithm is an extension of QDDS in a double Dirac delta potential well setup and uses a Chebyshev map driven solution update. The evolutionary behavior of QDDS is simple to follow from an intuitive point of view and guides the particle set towards lower energy configurations under the influence of a spatially co-located attractive double delta potential. The current gradient-dependent formulation is susceptible to getting trapped in suboptimal results because of the use of a gradient descent scheme in the $\delta_t$ computation phase. However, the algorithm is expensive in terms of time complexity because of a numerical approximation of $r_t$ from $\delta_t$ in the transcendental Eq. (25), as also outlined in Algorithm 1. As outlined in [1], the impact of cognition and social attractors, initial tessellation configurations, multi-scale topological communication schemes and correction (update) processes need to be studied to provide more insightful comments into the optimization of the workflow itself, specifically the stagnation issue and the high time complexity. In summary, the use of additional chaotic sequences in the heuristic evolution of QDDS based on this commonly used approximation abstraction from quantum physics remains to be further explored in light of the promising results obtained on some problems as highlighted in this study. Further, the snowball effect on the dynamics due to the selection of varying number of agents and selective communication among them over a user-defined number of generations is a thrust area gaining prominence as demonstrated in recent studies [28-29]. As we continue to further our understanding of how emergent properties arise out of simple, local-level interactions at the lowest hierarchical levels, we may expect the evolutionary computation community to increasingly consider scale-free interactions among atomic agents on top of the existing, already rich body of research on biomimicry. The proposed paradigm is well-suited for application in single-objective unimodal/multimodal optimization problems such as those discussed in [8,13-14,30] along the lines of digital filtering, fuzzy-clustering, scheduling, routing etc. The QDDS and subsequently C-QDDS approaches build on a growing corpus of algorithms hybridizing quantum swarm intelligence and global optimization and adds to the existing collection of nature-inspired optimization techniques.

**Acknowledgments:** This work was made possible by the financial and computing support by the Vanderbilt University Department of EECS.

**Author Contributions:** S.S. put forward the structure and organization of the article and created the content in all sections. S.S. ran the experiments and performed testing and convergence analyses and commented on the chaotic behavior of the algorithm. S.B. carried out precision testing and trajectory tracking analyses. Both S.S. and S.B. contributed to the final version of the article. R.A.P. II commented on the mathematical nature of the chaotic processes and provided critical analyses of assumptions in an advisory capacity. All authors approve of the final version of the article.

**Conflict of Interests:** The authors declare no conflict of interests.